\documentclass[sigconf]{acmart}
\usepackage{verbatim}
\usepackage{multirow}
\usepackage{caption}
\usepackage{graphicx}
\usepackage{ulem}
\usepackage{balance}

\AtBeginDocument{%
  \providecommand\BibTeX{{%
    \normalfont B\kern-0.5em{\scshape i\kern-0.25em b}\kern-0.8em\TeX}}}

\setcopyright{acmcopyright}
\copyrightyear{2021}
\acmYear{2021}
\setcopyright{acmlicensed}\acmConference[MM '21]{Proceedings of the 29th
ACM International Conference on Multimedia}{October 20--24, 2021}{Virtual
Event, China}
\acmBooktitle{Proceedings of the 29th ACM International Conference on
Multimedia (MM '21), October 20--24, 2021, Virtual Event, China}
\acmPrice{15.00}
\acmDOI{10.1145/3474085.3475648}
\acmISBN{978-1-4503-8651-7/21/10}

\begin{document}
\fancyhead{} 
\title{Knowledge Perceived Multi-modal Pretraining in E-commerce}

\author{Yushan Zhu\textsuperscript{*}}
\orcid{xxx}
\affiliation{%
  \institution{Zhejiang University}
  \streetaddress{P.O. Box 1212}
  \city{Hangzhou}
  \state{Zhejiang}
  \country{China}
}
\email{yushanzhu@zju.edu.cn}

\author{Huaixiao Tou\textsuperscript{*}}
\affiliation{%
  \institution{Alibaba Group}
  \city{Hangzhou}
  \state{Zhejiang}
  \country{China}}
\email{huaixiao.thx@alibaba-inc.com}

\author{Wen Zhang\textsuperscript{*}}
\affiliation{%
  \institution{Zhejiang University}
  \city{Hangzhou}
  \state{Zhejiang}
  \country{China}
}
\email{wenzhang2015@zju.edu.cn}

\author{Ganqiang Ye}
\affiliation{%
 \institution{Zhejiang University}
  \city{Hangzhou}
  \state{Zhejiang}
  \country{China}}
\email{yeganqiang@zju.edu.cn}

\author{Hui Chen}
\affiliation{%
  \institution{Alibaba Group}
  \streetaddress{30 Shuangqing Rd}
  \city{Hangzhou}
  \state{Zhejiang}
  \country{China}}
  \email{weidu.ch@alibaba-inc.com}

\author{Ningyu Zhang}
\affiliation{%
  \institution{Zhejiang University}
  \city{Hangzhou}
  \state{Zhejiang}
  \country{China}}
\email{zhangningyu@zju.edu.cn}

\author{Huajun Chen \textsuperscript{\textsection}}
\affiliation{%
  \institution{College of Computer Science \\ Hangzhou Innovation Center \\ Zhejiang University}
  \city{Hangzhou}
  \state{Zhejiang}
  \country{China}}
\email{huajunsir@zju.edu.cn}

\renewcommand{\shortauthors}{Zhu, et al.}

\begin{abstract}
  In this paper, we address multi-modal pretraining of product data in the field of E-commerce. Current multi-modal pretraining methods proposed for image and text modalities lack robustness in the face of modality-missing and modality-noise, which are two pervasive problems of multi-modal product data in real E-commerce scenarios. To this end, we propose a novel method, K3M, which introduces knowledge modality in multi-modal pretraining to correct the noise and supplement the missing of image and text modalities. The modal-encoding layer extracts the features of each modality. The modal-interaction layer is capable of effectively modeling the interaction of multiple modalities, where an initial-interactive feature fusion model is designed to maintain the independence of image modality and text modality, and a structure aggregation module is designed to fuse the information of image, text, and knowledge modalities. We pretrain K3M with three pretraining tasks, including masked object modeling (MOM), masked language modeling (MLM), and link prediction modeling (LPM). Experimental results on a real-world E-commerce dataset and a series of product-based downstream tasks demonstrate that K3M achieves significant improvements in performances than the baseline and state-of-the-art methods when modality-noise or modality-missing exists. 
\end{abstract}




\maketitle

\begingroup\renewcommand\thefootnote{*}
\footnotetext{Equal contribution.}
\begingroup\renewcommand\thefootnote{\textsection}
\footnotetext{Corresponding author.}

\section{Introduction}
The emergence of E-commerce has greatly facilitated people's lives. And there are a large number of product-based application tasks in the E-commerce scenario, such as item classification~\cite{DBLP:journals/kybernetes/WakilAGLR20,DBLP:conf/recsys/ZhaoCCJBBC18}, product alignment~\cite{DBLP:conf/www/SteinSW19}, recommendation system~\cite{DBLP:conf/recsys/ZhaoCCJBBC18,DBLP:conf/www/NeveP20,DBLP:conf/icde/WongFZVCZHCZC21} and so on. As shown in Figure~\ref{fig_intro}, there are usually images, titles, and structure knowledge of products, which is a typical multi-modal scenario.

\begin{figure}
\vspace{-0.2cm}
\setlength{\abovecaptionskip}{0.cm}
    \centering
    \includegraphics[width=0.49\textwidth]{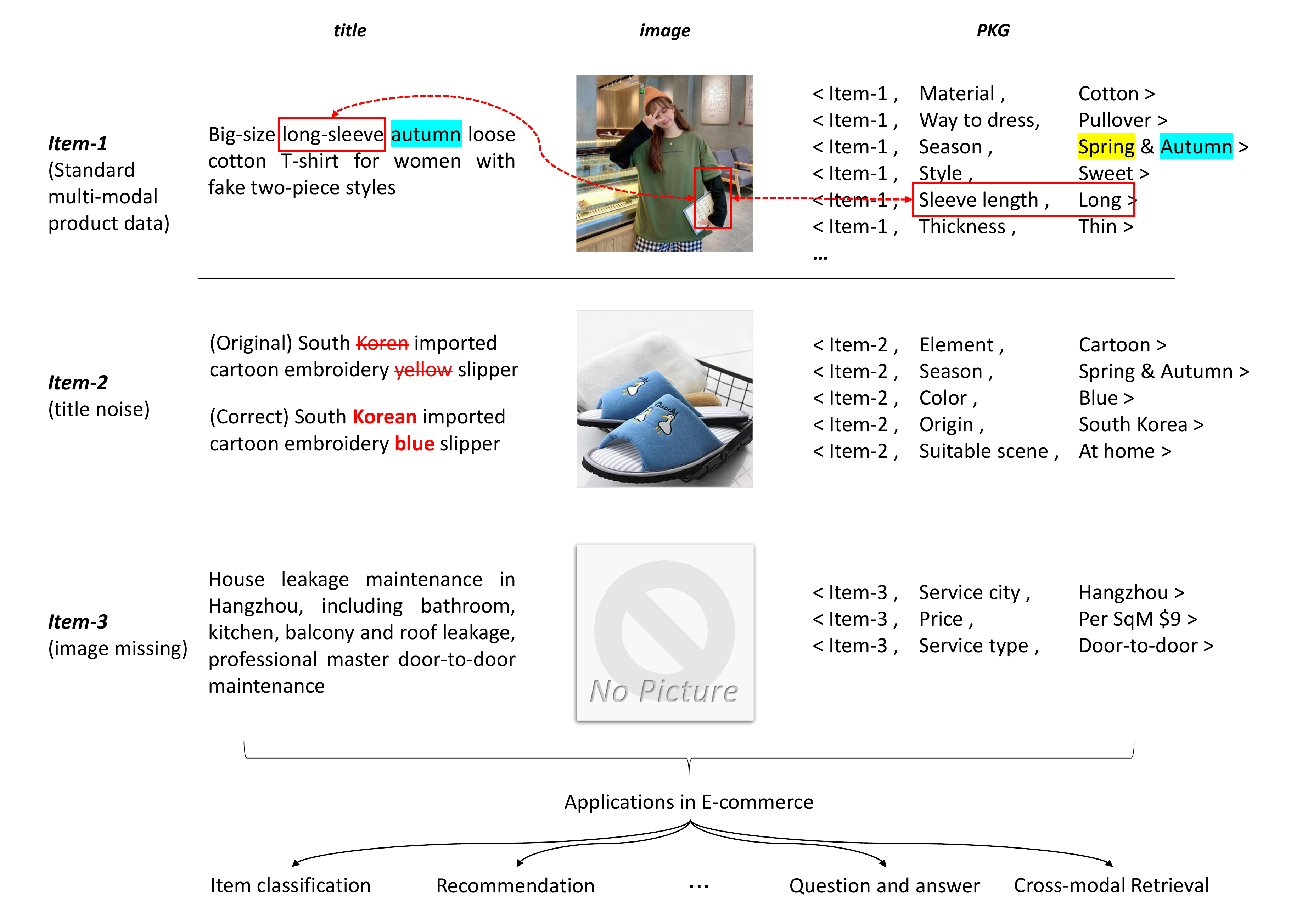}
    \caption{Examples of multi-modal data of products. Each item has a title, an image, and a PKG describing the objective properties of the product by triples as <\textit{item}, \textit{property}, \textit{value}>.}
    \label{fig_intro}
    \vspace{-0.4cm}
\end{figure}

Recently, multi-modal pretraining has attracted wide attention  ~\cite{16DBLP:conf/iccv/SunMV0S19,17DBLP:journals/corr/abs-1908-03557,18DBLP:conf/aaai/LiDFGJ20,19DBLP:conf/iclr/SuZCLLWD20,20DBLP:conf/emnlp/TanB19,21DBLP:conf/sigir/GaoJCQLWHW20,22DBLP:conf/nips/LuBPL19,36DBLP:journals/corr/abs-2003-13198,23DBLP:conf/aaai/ZhouPZHCG20, DBLP:conf/emnlp/AlbertiLCR19}, and these methods are dedicated to mining the association between information of image (or video) modality and text modality. Considering a wide range of downstream E-commerce applications, we focus on the pretraining of multi-modal data of products. However, applying these multi-modal pretraining methods directly to E-commerce scenarios will cause problems, because modality-missing and modality-noise are two challenges in the E-commerce scene, which will seriously reduce the performance of multi-modal information learning~\cite{24DBLP:conf/mm/YangWY20}. In a real E-commerce scenario, some sellers do not upload the product image (or title) to the platform, and some sellers provide the product image (or title) without accurate themes or semantics so that they are particularly puzzling. \textit{Item-2} and \textit{Item-3} in Figure~\ref{fig_intro} respectively shows an example of modality-noise and modality-missing in our scene. 

To solve this problem, we introduce Product Knowledge Graph (PKG)\cite{DBLP:conf/icde/ZhangWYWZC21} into consideration and regard it as a new modality called knowledge modality. As shown in Figure~\ref{fig_intro}, PKG contains triples in the form of <$h, r, t$>. For example, <\textit{Item-1}, \textit{Material}, \textit{Cotton}> represents that the material of \textit{Item-1} is cotton. We introduce PKG mainly for two reasons: (1) PKG has high quality. PKG describes the objective properties of the product, which is structured and easy to manage, and maintenance and standardization work are usually done for PKG. So PKG is relatively clean and credible. (2) Information contained in PKG and other modalities overlap each other. Take \textit{Item-1} in Figure~\ref{fig_intro} as an example, on the one hand, the image, title and PKG all tell that \textit{Item-1} is a long-sleeve T-shirt. On the other hand, PKG shows that this long-sleeve T-shirt is not only suitable for autumn, but also suitable for spring, which can't be known from the image or title. Thus PKG could correct or supplement other modalities when modality-noise or modality-missing exists. Therefore, for the pretraining of product data, we consider the information of three modalities: image modality (product image), text modality (product title), and knowledge modality (PKG). 

In this paper, we propose a novel \textbf{K}nowledge perceived \textbf{M}ulti-\textbf{M}odal pretraining \textbf{M}ethod in E-commerce application, named \textbf{K3M}. In particular, K3M learns the multi-modal information of products in 2 steps: (1) encoding the individual information of each modality, and (2) modeling the interaction between modalities.  
When encoding the individual information of each modality, for image modality, a Transformer-based image encoder is used to extract image initial features; for text modality, a Transformer-based text encoder is used to extract text initial features; for knowledge modality, the same text encoder is used to extract the surface form features of relations and tail entities of triples in PKG.

When modeling the interaction between modalities, there are two processes. The first is the interaction between text modality and image modality as did in previous work~\cite{22DBLP:conf/nips/LuBPL19}, and the second is the interaction between knowledge modality and the other two. In the first process, interactive features of image and text modalities are learned based on their initial features through co-attention Transformer~\cite{22DBLP:conf/nips/LuBPL19}. And to remain the independence of individual modality, we propose to fuse the initial features of image and text modalities with their interactive features by an initial-interactive feature fusion module. In the second process, the interaction result of image and text modalities is used to initialize the representation of the target product entity, which is the head entity of triples in PKG, and the surface form features of relations and tail entities are used as their initial representations. Then the information of entities and relations is propagated and aggregated on the target product entity through a structure aggregation module. Finally, the knowledge-guided representation of product entities can be used for various downstream tasks. 

The pretraining tasks for image modality, text modality, and knowledge modality are masked object modeling, masked language modeling, and link prediction modeling, respectively.

The experimental results on several downstream tasks show that our K3M is more robust than current multi-modal pretraining methods in modeling entity. Our main contributions are as follows:
\begin{itemize}
\item We introduce structured knowledge of PKG into multi-modal pretraining in E-commerce, which can correct or weaken the modality-noise and modality-missing problems in large-scale multi-modal datasets.
\item We propose a novel multi-modal pretraining method, K3M. In K3M, we fuse the initial features of image and text modalities with their interactive features to further improve the model performance.
\item Experiments on a real-world E-commerce dataset show the powerful ability of K3M in many downstream tasks. Our code and dataset is available at https://github.com/YushanZhu/K3M.
\end{itemize}

\section{Related Work}

\subsection{Multi-Modal Pretraining}
The success applications of pretraining technique in the field of computer vision (CV), such as VGG~\cite{30DBLP:journals/corr/SimonyanZ14a}, Google Inception~\cite{DBLP:conf/cvpr/SzegedyVISW16} and ResNet~\cite{DBLP:conf/cvpr/XieGDTH17}, and natural language processing (NLP), such as BERT~\cite{25DBLP:conf/naacl/DevlinCLT19}, XLNet~\cite{26DBLP:conf/nips/YangDYCSL19} and GPT-3~\cite{DBLP:conf/nips/BrownMRSKDNSSAA20}, have inspired the development of multi-modal pretraining. Recently a series of multi-modal pretraining methods have been proposed, where information from different modalities complements each other.

VideoBERT~\cite{16DBLP:conf/iccv/SunMV0S19} is the first work of multi-modal pretraining, which trains a large number of unlabeled video-text pairs through BERT. At present, there are two main architectures of multi-modal pretraining models for image and text.
B2T2~\cite{DBLP:conf/emnlp/AlbertiLCR19}, VisualBERT~\cite{17DBLP:journals/corr/abs-1908-03557}, Unicoder-VL~\cite{18DBLP:conf/aaai/LiDFGJ20}, VL-BERT~\cite{19DBLP:conf/iclr/SuZCLLWD20} and UNITER~\cite{35DBLP:conf/eccv/ChenLYK0G0020} propose the single-stream architecture, where a single Transformer is applied
to both images and text. On the other hand, LXMERT~\cite{20DBLP:conf/emnlp/TanB19}, ViLBERT~\cite{22DBLP:conf/nips/LuBPL19} and FashionBERT~\cite{21DBLP:conf/sigir/GaoJCQLWHW20} introduce the two-stream architecture, where the features of image and text are first extracted independently, and then a more complex mechanism named co-attention is applied to complete their interaction. To further boost the performance,
VLP~\cite{23DBLP:conf/aaai/ZhouPZHCG20} applies a shared multi-layer Transformer for encoding and decoding, which is used for both image captioning and VQA. Based on the single-stream architecture, InterBERT~\cite{36DBLP:journals/corr/abs-2003-13198} adds two streams of separate
Transformer to the output of the single-stream model to capture the modal independence. 
These multi-modal pretraining methods cannot solve the problem of modality-missing and modality-noise. Compared with the previous work, K3M has several significant differences. Our proposed model architecture can effectively utilize the structured knowledge to improve the robustness of the model against modality-missing and modality-noise. In addition, we propose to fuse modal initial features and interactive features to retain the independence of text and image modalities, which makes the model more effective.

\subsection{KG-enhanced pretraining models}
Recently, more and more researchers pay attention to the combination of knowledge graph (KG) and pretrained language model (PLM) to enable PLMs to reach better performance. 

K-BERT \cite{DBLP:conf/aaai/LiuZ0WJD020} injects triples into a sentence to generate a unified knowledge-rich language representation. 
ERNIE~\cite{DBLP:conf/acl/ZhangHLJSL19} integrates entity representations from the knowledge module into the semantic module to represent heterogeneous information of tokens and entities into a united feature space. 
KEPLER~\cite{DBLP:journals/tacl/WangGZZLLT21} encodes textual descriptions for entities as text embeddings and treats the description embeddings as entity embeddings.
KnowBert~\cite{DBLP:conf/emnlp/PetersNLSJSS19} uses an integrated entity linker to generate knowledge enhanced entity-span representations via a form of word-to-entity attention.
K-Adapter \cite{wang2020K_Adapter} injects factual knowledge and linguistic knowledge into RoBERTa with a neural adapter for each kind of infused knowledge. 
DKPLM \cite{su2020DKPLM} could dynamically select and embed knowledge according to the textual context for PLMs, with the awareness of both global and local KG information.
JAKET \cite{yu2020JAKET} proposes a joint pretraining framework, including the knowledge module to produce embeddings for entities to generate context-aware embeddings in the graph. What's more, KALM~\cite{rosset2020KALM}, ProQA~\cite{DBLP:conf/eacl/XiongWW21}, LIBERT~\cite{lauscher2019informing} and other researchers explore the fusion experiment with knowledge graphs and PLMs in different application tasks.

However, the current KG-enhanced pretraining models only aim at single modality, especially text modality. To the best of our knowledge, this is the first work that incorporates knowledge graph into multi-modal pretraining.

\section{Methods}

\begin{figure*}
\setlength{\abovecaptionskip}{0.05cm}
    \centering
    \includegraphics[width=1.17\textwidth]{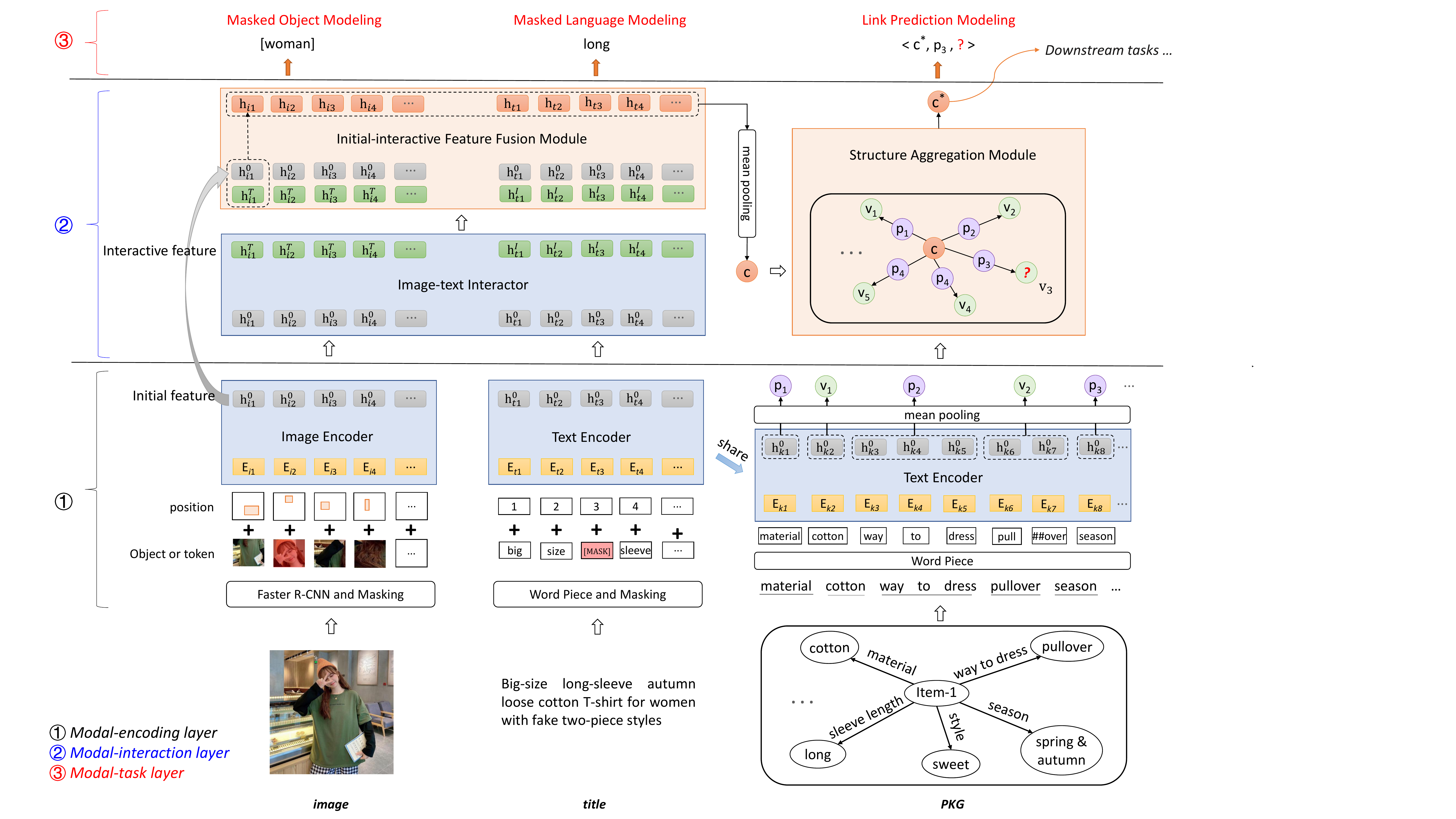}
    \caption{Model framework of K3M.}
    \label{fig:model_framework}
    \vspace{-0.3cm}
\end{figure*}

In this section, we will describe how K3M jointly models the information of text, image and knowledge modality. Given a set of product data $\mathcal{D}=\{\mathcal{C}, \mathcal{I}, \mathcal{T}, \mathcal{K}\}$, where $\mathcal{C}$ is a set of products, $\mathcal{I}$ is a set of product images, $\mathcal{T}$ is a set of product titles,  $\mathcal{K} = \{ \mathcal{E}, \mathcal{R}, \mathcal{TR} \}$ is PKG where $\mathcal{E}$, $\mathcal{R}$ and $\mathcal{TR}$ are the set of entities, relations and triples and $\mathcal{TR} = \{ <h,r,t> | h \in \mathcal{E}, r \in \mathcal{R}, t \in \mathcal{E}\}$. For each item $e_c \in \mathcal{C}$, 
it has an product image $i_c \in \mathcal{I}$, a product title $t_c \in \mathcal{T}$ and a set of triples from $\mathcal{K}$ that are related to $e_c$, namely $\mathcal{TR}_c=\{<e_c, property, value> | e_c \in \mathcal{E}, property \in \mathcal{R}, value \in \mathcal{E} \} \subset \mathcal{TR}$. Our target is to learn a model $M(i_c,t_c,\mathcal{TR}_c)$ to learn a good cross-modal representation for $e_c \in \mathcal{C}$, represented as $c^\ast$.

Our model consists of three layers, as shown in Figure~\ref{fig:model_framework}. The first layer named modal-encoding layer aims to separately encode the individual information of each modality. The second layer named modal-interaction layer aims to model the interaction between different modalities. The third layer is modal-task layer, and there are different pretraining tasks for different modalities. We (1) first describe how to encode image initial features, text initial features, and surface form features of knowledge in modal-encoding layer. (2) And then we demonstrate the two processes of modeling the interaction between modalities in modal-interaction layer. (3) Finally, we describe the three pretraining tasks in modal-task layer.

\subsection{Modal-encoding layer}
\subsubsection{Image initial features.}
Following ~\cite{22DBLP:conf/nips/LuBPL19,20DBLP:conf/emnlp/TanB19,19DBLP:conf/iclr/SuZCLLWD20,17DBLP:journals/corr/abs-1908-03557,36DBLP:journals/corr/abs-2003-13198}, we transform the product image $i_c$ of a given item $e_c$, a matrix of pixels, into a object sequence through the object detection model. Specifically, following ViLBERT~\cite{22DBLP:conf/nips/LuBPL19}, we apply Faster R-CNN~\cite{40DBLP:conf/nips/RenHGS15} to 
detect a series of objects (RoIs regions of interest) from $i_c$, and the bounding boxes of these objects are used as their positional information.
15\% of the objects are randomly masked as in~\cite{22DBLP:conf/nips/LuBPL19}. Then as shown in Figure~\ref{fig:model_framework}, the sum of the object embedding and positional embedding $[E_{i1},E_{i2},...,E_{iM_1}]$ is input to a Transformer-based image encoder, and the image encoder outputs the image initial features $[h^0_{i1},h^0_{i2},...,h^0_{iM_1}]$, where $M_1$ is the maximum object sequence length.

\subsubsection{Text initial features.} 
Following BERT~\cite{25DBLP:conf/naacl/DevlinCLT19}, the product title $t_c$ of $e_c$ is first tokenized into a token sequence by WordPieces~\cite{38DBLP:journals/corr/WuSCLNMKCGMKSJL16}, and 15\% of the tokens are randomly masked. Then as shown in Figure~\ref{fig:model_framework}, the sum of the token embedding and positional embedding $[E_{t1},E_{t2},...,E_{tM_2}]$ is input to a Transformer-based text encoder, and the text encoder outputs the text initial features $[h^0_{t1},h^0_{t2},...,h^0_{tM_2}]$, where $M_2$ is the maximum token sequence length.

\subsubsection{Surface form features of knowledge.} In this step, we obtain the surface form features of the relations and the tail entities of the triples <$e_c$, $property_x$, $value_x$> in $\mathcal{TR}_c$, where $x=1,...,X_c$ and $X_c$ is the number of triples in $\mathcal{TR}_c$. Here we do not consider the head entity because its surface form has no semantics as shown in Figure~\ref{fig_intro}. To make full use of the contextual information, we first stitch all relations and tail entities of the triples in $\mathcal{TR}_c$ together into a long knowledge text like ``$property_1$ $value_1$ $property_2$ $value_2$ $property_3$ $...$'' (for example, the knowledge text of \textit{Item-1} in Figure~\ref{fig_intro} is ``material cotton way to dress pullover season ...''), and then tokenize it into a token sequence according to WordPieces. After that, the same text encoder for extracting text initial features in section 3.1.2 is used to encode the token sequence of knowledge text. As shown in Figure~\ref{fig:model_framework}, the text encoder outputs $[h^0_{k1},h^0_{k2},...,h^0_{kM_3}]$ based on the input embedding $[E_{k1},E_{k2},...,E_{kM_3}]$, where $M_3$ is the maximum token sequence length for knowledge text. Finally, we calculate the surface form features of each relation and tail entity as the mean-pooling value of the last hidden layer state of their corresponding tokens (one relation or tail entity may be tokenized into several tokens as shown in Figure~\ref{fig:model_framework}), denoted as $p_x$ and $v_x$, where $x=1,...,X_c$ and $X_c$ is the number of triples in $\mathcal{TR}_c$.

\subsection{Modal-interaction layer}
In this layer, there are two process to model the modal interaction. The first is the interaction between image modality and text modality, and the second is the interaction between knowledge modality and the other two. We will separately introduce the two processes.

\subsubsection{Interaction between image modality and text modality}
First, an image-text interactor, applying the co-attention Transformer~\cite{22DBLP:conf/nips/LuBPL19}, takes the image initial features $[h^0_{t1},h^0_{t2},...,h^0_{tM_1}]$ and text initial features $[h^0_{i1},h^0_{i2},...,h^0_{iM_2}]$ as input. Specifically, in co-attention Transformer, the ``key'' and ``value'' in attention blocks of each modality are passed to the attention blocks of the other modality, performing image-conditioned text attention and text-conditioned image attention. After that, the image-text interactor produces interactive features $[h^T_{i1},h^T_{i2},...,h^T_{iM_1}]$ for image conditioned on the text and $[h^I_{t1},h^I_{t2},...,h^I_{tM_2}]$ for text conditioned on the image. 

However, when learning the modal interactive features through co-attention Transformer, the independence of individual modality is ignored~\cite{36DBLP:journals/corr/abs-2003-13198}. When a modality has noise or missing, the modal interaction will have a negative impact on the other modality, thereby destroying the modal interactive features. Thus, it is necessary to maintain the independence of individual modality. To solve this problem, we propose to retain the image initial features and text initial feature learned in the modal-encoding layer, and design an \textbf{initial-interactive feature fusion module (IFFM)} to fuse the initial features and the interactive features of image and text modalities. IFFM takes the initial feature and interactive feature of an object (or a token) as input, and fuse the two feature vectors into an output vector, expressed as:
\begin{equation}
\setlength{\abovedisplayskip}{3pt}
\setlength{\belowdisplayskip}{3pt}
\label{eq_fusion}
\begin{split}
    h_{ta}&=fusion(h^0_{ta},h^I_{ta}),(a=1,2,...,M_1),\\
    h_{ib}&=fusion(h^0_{ib},h^T_{ib}),(b=1,2,...,M_2),
\end{split}
\end{equation}
where function $fusion(\cdot)$ is fusion algorithm, and there are three fusion algorithms in K3M: (1) Mean: calculating the mean of two input vectors, the model is denoted as ``K3M(mean)''. (2) Soft-Sampling: an advanced sampling method proposed for feature fusion in~\cite{42DBLP:conf/cvpr/ChenRMLWMT19}, the model is denoted as ``K3M(soft-spl)''. (3) Hard-Sampling: another advanced sampling method proposed for feature fusion in~\cite{42DBLP:conf/cvpr/ChenRMLWMT19}, the model is denoted as ``K3M(hard-spl)''.

\subsubsection{Interaction between knowledge modality and the other two modalities}

First, the interaction result of image and text modalities is used to initialize the representation of item $e_c$, which is the head entity of the triples in $\mathcal{TR}_c$. We calculate the initial representation of $e_c$ as the mean-pooling value of all output of IFFM:
\begin{equation}
\setlength{\abovedisplayskip}{3pt}
\setlength{\belowdisplayskip}{3pt}
\label{eq_initial_c}
c=mean\_pooling(h_{t1},...,h_{tM_1},W_0h_{i1},...,W_0h_{iM_2})
\end{equation}
where $W_0$ is a linear transformation matrix to convert all vectors to the same dimension. So the representations of head entity, relation, tail entity of triple <$e_c$, $property_x$, $value_x$> ($x=1,...,X_c$) can be respectively initialized as $c$, $p_x$, and $v_x$, where $p_x$ and $v_x$ are surface features of the relations and tail entities learned in modal-encoding layer.

Inspired by the idea of~\cite{41DBLP:conf/acl/NathaniCSK19}, an improvement of GAT~\cite{45DBLP:conf/iclr/VelickovicCCRLB18} and aiming to capture both entity and relation features in any given entity's neighborhood, we design a \textbf{structure aggregation module} to propagate and aggregate the information of entities and relations, so as to fuse the information of image, text and knowledge modalities. Specifically, the representation of each triple <$e_c, property_x, value_x$> is first learned by:
\begin{equation}
\setlength{\abovedisplayskip}{3pt}
\setlength{\belowdisplayskip}{3pt}
\label{eq_tri_emb}
    t_{x}=W_1[c||p_x||v_x],
\end{equation}
where $W_1$ is a linear transformation matrix. Then, the importance of the triple is denoted by the $LeakyRelu$ non-linearity as:
\begin{equation}
\setlength{\abovedisplayskip}{3pt}
\setlength{\belowdisplayskip}{3pt}
\label{eq_tri_imp}
    b_{x}=LeakyReLU(W_2t_{x}),
\end{equation}
where $W_2$ is a linear weight matrix. And the attention value of each triple is obtained by applying $Softmax$:
\begin{equation}
\setlength{\abovedisplayskip}{3pt}
\setlength{\belowdisplayskip}{3pt}
\label{eq_tri_att}
    a_{x}  =Softmax(b_{x})
 =\frac{exp(b_{x})}{\sum_{i=1}^{X_c}exp(b_{i})}.
\end{equation}
Finally, the final representation of item $e_c$ is obtained by adding its initial representation $c$ and the sum of the representations of all triples in $\mathcal{TR}_c$ weighted by their attention values as:
\begin{equation}
\setlength{\abovedisplayskip}{3pt}
\setlength{\belowdisplayskip}{3pt}
\label{eq_tri_repnew}
c^\ast=W_3c+ \sigma \Bigl(\frac{1}{M_h}\sum_{m=1}^{M_h}\sum_{x=1}^{X_c}a_{x}^mt_{x}^m\Bigr),
\end{equation}
where $W_3$ is a weight matrix, $\sigma(\cdot)$ is the activate function, $M_h$ is the number of attention heads, and $X_c$ is the number of triples in $\mathcal{TR}_c$.

\subsection{Modal-task layer} In this layer, we exploit different pretraining tasks for the three modalities. They are masked language modeling (MLM) for text modality, masked object modeling (MOM) for image modality, and link prediction modeling (LPM) for knowledge modality.

\subsubsection{Masked Language Modeling (MLM)}
This task is the same as the MLM task in BERT pretraining, whose objective is to predict the masked tokens. The training minimizes the cross-entropy loss:
\begin{equation}
\setlength{\abovedisplayskip}{3pt}
\setlength{\belowdisplayskip}{3pt}
    l_{MLM}=-E_{t_c\sim \mathcal{T}}logP(tok_m|tok_{\widehat{m}}),
\end{equation}
where $tok_m$ refers to the masked tokens, and $tok_{\widehat{m}}$ refers to the token sequence in which $tok_m$ has been masked.

\subsubsection{Masked Object Modeling (MOM)}
Similar to MLM, the objective of MOM is to predict the categories of the masked objects in the image. The training minimizes the cross-entropy loss:
\begin{equation}
\setlength{\abovedisplayskip}{3pt}
\setlength{\belowdisplayskip}{3pt}
    l_{MOM}=-E_{i_c\sim \mathcal{I}}logP(obj_m|obj_{\widehat{m}}),
\end{equation}
where $obj_m$ refers to the masked objects, and $obj_{\widehat{m}}$ refers to the object sequence in which $obj_m$ has been masked.

\subsubsection{Link Prediction Modeling (LPM)}
The goal of this task is to evaluate the credibility of a given triple. Following the translation-based KG embedding method TransE~\cite{39DBLP:conf/nips/BordesUGWY13}, which assumes that if <$h,r,t$> is a true triple, the representation vector of $h$ plus the representation vector of $r$ should be equal to the representation vector of $t$, we define the score of triple <$e_c$, $property_x$, $value_x$> as $S^x_c=||c^\ast+p_x-v_x||_1$, $x=1,...,X_c$. The objective of LPM is to make the score lower for the correct triples while higher for the wrong ones. The training minimizes the margin-loss:
\begin{equation}
\setlength{\abovedisplayskip}{3pt}
\setlength{\belowdisplayskip}{3pt}
\label{eq_los_tri}
    l_{LPM}=E_{\mathcal{TR}_c\sim \mathcal{K}}\frac{1}{X_c}\sum_{x=1}^{X_c} max\{S_c^x-\widetilde{S_{c}^{x}}+\gamma,0\},
\end{equation}
where $\gamma$ is a margin hyper-parameter, $S_c^x$ is the score of the positive triple <$e_c$, $property_x$, $value_x$>,  and $\widetilde{S_c^x}$ is the score of the negative triple <$e_c'$, $property_x$, $value_x$> or <$e_c$, $property_x$, $value_x'$> generated by randomly replacing the head entity $e_c$ or tail entity $value_x$ with any other entity in $\mathcal{E}$.

And the final pretraining loss of K3M is the sum of the losses of the above three tasks:
\begin{equation}
\setlength{\abovedisplayskip}{3pt}
\setlength{\belowdisplayskip}{3pt}
L_{pre}=l_{MLM}+l_{MOM}+l_{LPM}.
\end{equation}

\section{Experiments}
\subsection{Pretraining}
\subsubsection{Dataset.} 

Our K3M is trained and validated on millions of items, $40132418$, from Taobao, where each item contains a title, an image and a set of triples related to it from PKG. The statistics of pretraining dataset are shown in Table~\ref{T-data}.
\vspace{-0.2cm}
\begin{table}[!h]
\setlength{\abovecaptionskip}{0.15cm}
\caption{Pretraining data statistics. }
\centering
\setlength\tabcolsep{4pt}   
\resizebox{0.4\textwidth}{!}{
\begin{tabular}{lcccc}
\toprule
        & \# Train    & \# Valid & \# Test & Total       \\[-1pt] 
\hline
Items   & 39,966,300  & 33,704   & 132,414 & 40,132,418  \\[-1pt]
Triples & 287,445,622 & 245,162  & 955,512 & 288,646,296 \\[-2pt] 
\bottomrule
\end{tabular}
}
\label{T-data}
\end{table}

Our K3M is pretrained on the train and the valid dataset, and we evaluate the pretrained K3M on the test dataset, which is used as the finetuning dataset for downstream tasks.

\subsubsection{Implementation details} 
We implement K3M with Pytorch in which three fusion algorithms, named ``K3M(mean)'', ``K3M(soft-spl)'' and ``K3M(hard-spl)'', are applied in the initial-interactive fusion module. 
More details are in Appendix.

\subsubsection{Pretraining of Baselines.}
We compare K3M with several image and text modality pretraining baselines:
a representative single-stream method VLBERT~\cite{19DBLP:conf/iclr/SuZCLLWD20}, and two two-stream methods ViLBERT~\cite{22DBLP:conf/nips/LuBPL19} and LXMERT~\cite{20DBLP:conf/emnlp/TanB19}.
Baselines are also pretrained on the Taobao data and initialized following their original papers.
We pretrain two types of models for baselines: 
(1) training with normal image and text modality which include ``ViLBERT'', ``LXMERT'' and  ``VLBERT'';
(2) training with image, text, and knowledge modality which include ``ViLBERT+PKG'', ``VLBERT+PKG'', and ``LXMERT+PKG'', where knowledge text from PKG are spliced behind title text as the text modality input. More details are in Appendix.



\subsection{Finetuning: Item Classification}

\begin{figure*}
\vspace{-0.2cm}
    \centering
    \setlength{\abovecaptionskip}{0.05cm}
    \includegraphics[width =0.95\textwidth]{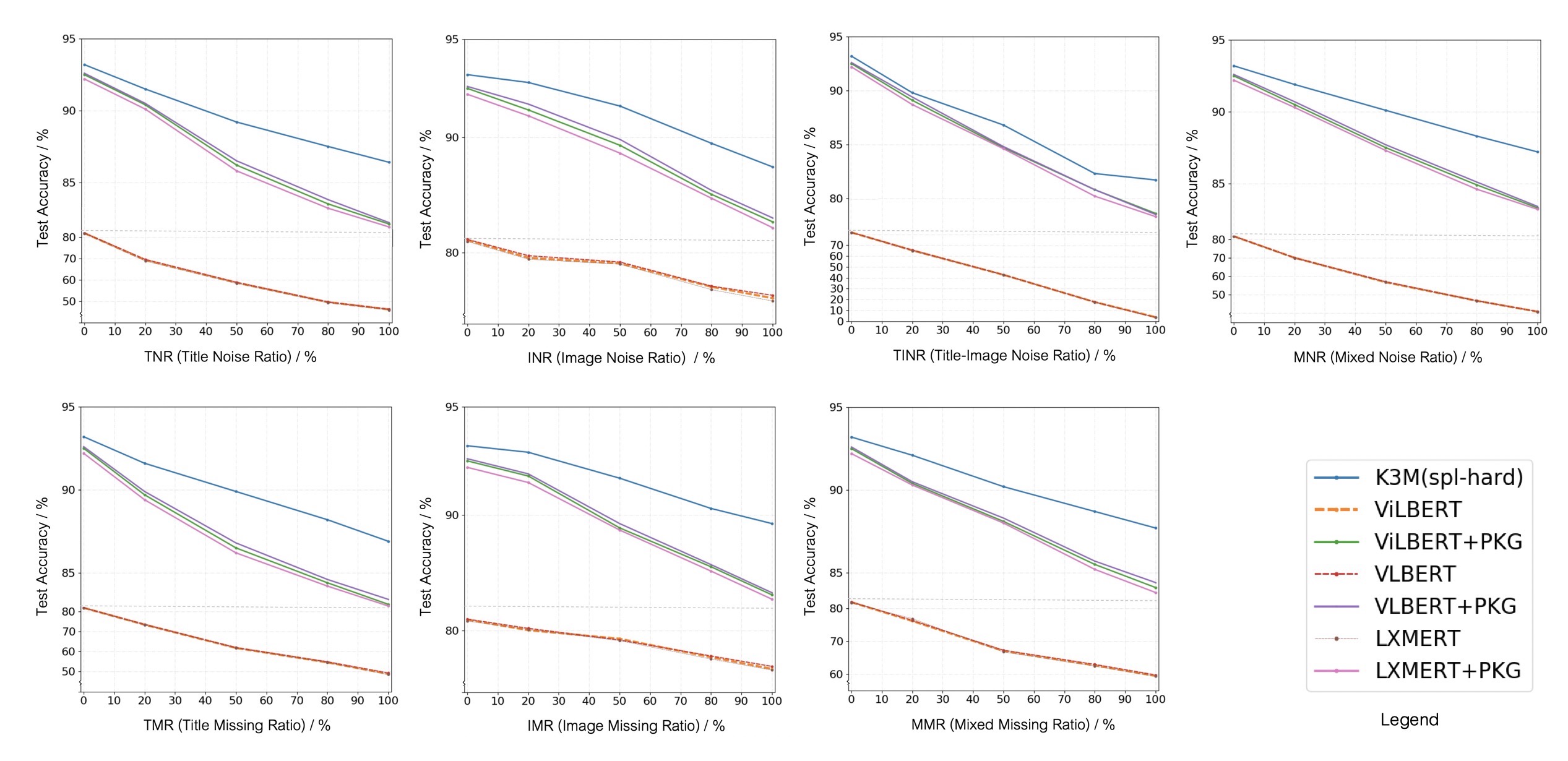}
    \caption{Test accuracy (\%) for item classification with various missing and noise settings.}
    \label{fig:PKGM_1}
    \vspace{-0.2cm}
\end{figure*}

\subsubsection{Task definition} 
Item classification is to assign given items to corresponding classes which is a common task for product management in E-commerce platform and could be regarded as a multi-class classification task as 
given a set of items $\mathcal{C}$ and a set of classes $\mathcal{CLS}$, the target is to train a mapping function  $f: \mathcal{C} \longmapsto \mathcal{CLS}$. 

\subsubsection{Model} 
\textbf{K3M.} For a given item $e_{ci} \in \mathcal{C}$, which contains an image $i_c$, a title $t_c$, and a set of triples $\mathcal{TR}_c$, we get its representation $c_i ^\ast =M(i_c,t_c,\mathcal{TR}_c)$ and feed it into a full connection layer:
\begin{equation}
\setlength{\abovedisplayskip}{3pt}
\setlength{\belowdisplayskip}{3pt}
\label{equ_itemcls1}
    p_i=\sigma(Wc_i^\ast + \beta),
\end{equation}
where $W\in \mathbb R^{d\times |\mathcal{CLS}|}$ is a weighted matrix, $d$ is the dimension of $c_i^ \ast$, $\beta$ is bias vector, $p_i=[p_{i1},p_{i2},...,p_{i|\mathcal{CLS}|}]$ where $p_{ij}$ is the probability that item $e_{c_i}$ belongs to class $cls_j$, $j \in \{1,...,|\mathcal{CLS}| \}$. We finetune K3M with a cross-entropy loss:
\begin{equation}
\setlength{\abovedisplayskip}{3pt}
\setlength{\belowdisplayskip}{3pt}
\label{equ_itemcls2}
    L=-\frac{1}{|\mathcal{C}|}\sum_{i=1}^{|\mathcal{C}|}\sum_{j=1}^{|\mathcal{CLS}|} y_{ij} log(p_{ij}),
\end{equation}
where $y_{ij} = 1$ if $e_{ci}$ belongs to class $cls_j$, otherwise $y_{ij} = 0$.

\textbf{Baseline models.} Following  original papers, we compute the representation of $e_{ci}$ as the element-wise product between the last hidden states of $[IMG]$ and $[CLS]$ for ViLBERT, and as the last hidden state of $[CLS]$ for LXMERT and VLBERT. The following steps are the same as Equation~(\ref{equ_itemcls1}) and (\ref{equ_itemcls2}).

\subsubsection{Dataset} 
More details are in Appendix.

\subsubsection{Missing and noise of Dataset} For a dataset that contains $N$ items, we set 3 missing situations for product title and image:
\begin{itemize}
    \item Title-only missing  
ratio \textbf{TMR}=$\rho \%$ denotes that $\rho \%$ items do not have titles. 
\item Image-only missing ratio \textbf{IMR}=$\rho \%$ denotes that $\rho \%$ of all items do not have images.
\item Mixed missing ratio \textbf{MMR}=$\rho \%$ denotes that  $\rho \%$ items have different conditions of title-only and image-only missing.  As in~\cite{DBLP:conf/kdd/ChenZ20}, for each item class, we randomly sample $N \times \rho/2 \%$ items to remove their images, and sample $N \times \rho/2 \%$ items from the rest to remove their titles. 
\end{itemize}

And we set 4 noise situations for product title and image: 
\begin{itemize}
    \item Title-only noise \textbf{TNR}=$\rho \%$ denotes that $\rho \%$ items have titles that don't match them, which are created via replacing item titles with any other items' titles.
    \item Image-only noise ratio \textbf{INR}=$\rho \%$ denotes that $\rho \%$ of all items have images that don't match them, which are created by replacing item images with any other items' images. 
    \item Title-image noise ratio \textbf{TINR}=$\rho \%$ denotes that $\rho \%$ of all items have both images and titles that don't match them at the same time, which are created by randomly replacing both of their titles and images as introduced before. 
    \item Mixed noise. The mixed noise ratio \textbf{MNR}=$\rho \%$ denotes that a total of $\rho \%$ items have different conditions of title-only, image-only, and title-image noise. As in~\cite{DBLP:conf/kdd/ChenZ20}, for each item class, we randomly sample $N \times \rho/3 \%$ items to replace their images/titles,
    and randomly sample $N \times \rho/3 \%$ items from the rest to replace both their titles and images. 
\end{itemize}


Following~\cite{DBLP:conf/kdd/ChenZ20}, to make balanced datasets, we keep the number of items for each class and each missing or noise situation in train/dev/test dataset as $7:1:2$. The settings of modal missing and noise of datasets are the same for all downstream tasks.


\subsubsection{Result analysis}
Figure~\ref{fig:PKGM_1}
\footnote{In Figure~\ref{fig:PKGM_1} and \ref{fig:PKGM_2}, reults of K3M(hard-spl) are shown since it works the best. For more results of K3M(mean) and K3M(soft-spl), please refer to Table \ref{T-ablation}.} 
shows results of various models for item classification, from which we have the following observations:  
(1) Baseline models seriously lack robustness when modality-missing or modality-noise exists. For ``title-only missing'', performance of  ``ViLBERT'', ``LXMERT'' and ``VLBERT'' decreases on average $10.2\%$, $24.4\%$, $33.1\%$ , and $40.2\%$ as TMR increases to $20\%$, $50\%$, $80\%$, and $100\%$, compared with TMR=$0\%$.
(2) Text modality with missing and noise have greater impact on the performance than image modality. Comparing the ``title-only noise'' and ``image-only noise'' of the 3 baselines, the model performance decreases by between $15.1\%$ and $43.9\%$ as TNR increases, while between $2.8\%$ and $10.3\%$ as INR increases, which indicates that text information plays a more important role.
(3) The introduction of knowledge graph can significantly improve the problem of modality-missing and modality-noise. The experimental results of baselines with PKG are significantly better than those without PKG. For ``title-only missing'', on the basis of baselines without PKG, ``ViLBERT+PKG'', ``LXMERT+PKG'' and ``VLBERT+PKG'' achieve an average improvement of $13.0\%$, $22.2\%$, $39.9\%$, $54.4\%$ and $70.1\%$ when TMR increases from $0\%$ to $100\%$.
(4) Our method achieves state-of-the-art performance on these benchmarks. It further improves the results of ``ViLBERT+PKG'', ``LXMERT+PKG'' and ``VLBERT+PKG'' by between $0.6\%$ and $4.5\%$ on various modality-missing and modality-noise settings.

\begin{figure*}
\vspace{-0.2cm}
    \centering
    \setlength{\abovecaptionskip}{0.05cm}
    \includegraphics[width = 0.95\textwidth]{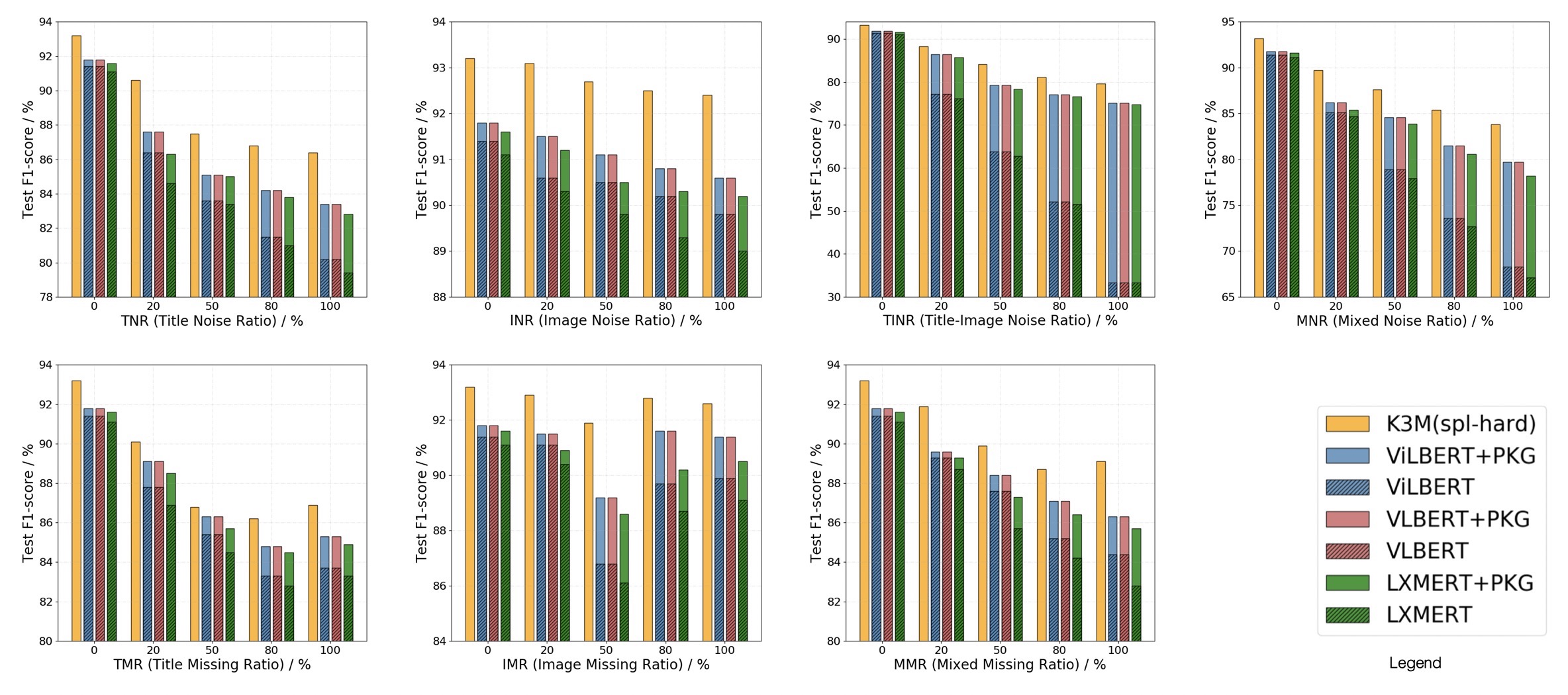}
    \caption{Test F1-score (\%) for product alignment with various missing and noise settings.}
    \label{fig:PKGM_2}
    \vspace{-0.2cm}
\end{figure*}


\subsection{Finetuning: Product Alignment}
\subsubsection{Task definition} 
Product alignment is to tell whether a given pair of items are aligned (referring to the same product). For example, there are many online shops selling IPhone 11 with White color and 128 GB capacity. They are different items on the platform, while from the perspective of the product, they refer to the same product. This task greatly helps daily business, such as recommending products to help the user compare their prices and after-sale services. 
It could be regarded as a binary classification task as given a set of item pairs $\mathcal{C}^p$ and a set $\mathcal{Y}=\{True,False \}$, the target is to train a mapping function $f: \mathcal{C}^p \longmapsto \mathcal{Y}$.

\subsubsection{Model} 
\textbf{K3M} For a given item pair $(e_{ci}^0,e_{ci}^1) \in \mathcal{C}^p$, we first get the representations of item $e_{ci}^0$ and $e_{ci}^1$ respectively, namely $c^\ast_{i0}=M(i_c^0,t_c^0,\mathcal{TR}_c^0)$ and $c^\ast_{i1}=M(i_c^1,t_c^1,\mathcal{TR}_c^1)$, which then are concatenated and fed into a full connection layer:
\begin{equation}
\setlength{\abovedisplayskip}{6pt}
\setlength{\belowdisplayskip}{3pt}
\label{equ_samepro1}
    p_i=\sigma(W[c^\ast_{i0}||c^\ast_{i1}] + \beta),
\end{equation}
where $W\in \mathbb R^{2d\times 2}$, $p_i=[p_{i0},p_{i1}]$ 
where $p_{i1}$ is the probability that $e_{ci}^0$ and $e_{ci}^1$ are aligned. We finetune K3M with a cross-entropy loss:
\begin{equation}
\setlength{\abovedisplayskip}{3pt}
\setlength{\belowdisplayskip}{3pt}
\label{equ_samepro2}
    L=-\frac{1}{|\mathcal{C}^p|}\sum_{i=1}^{|\mathcal{C}^p|} y_{i} log(p_{i1})+(1-y_{i})log(p_{i0}),
\end{equation}
where $y_{i}=1$ if $e_{ci}^0$ and $e_{ci}^1$ are aligned, otherwise $y_{i} = 0$.

\textbf{Baseline models} We calculate the item representations of $e_{ci}^0$ and $e_{ci}^1$ respectively in the same way as in item classification task. The following steps are the same as Equation~\ref{equ_samepro1} and \ref{equ_samepro2}.

\subsubsection{Dataset} 
More details are in Appendix.



\subsubsection{Result analysis.}

The evaluation metric of this task is F1-score. Figure~\ref{fig:PKGM_2} shows the test F1-score of product alignment task. In this task, we can have the similar observations as in the item classification task. In addition, for modality-missing, the model performance does not necessarily decrease as the missing ratio ncreases, but fluctuates: When the missing ratio (TMR, IMR and MMR) is $50\%$ or $80\%$, the model performance sometimes is even lower than when it is $100\%$. Actually, the essence of this task is to learn a model to evaluate the similarity of the multi-modal information of two items. Intuitively, when the two items of an aligned item pair lack titles or images at the same time, their information looks more similar than when one lacks title or image while the other lacks nothing.

\subsection{Finetuning: Multi-modal Question Answering}
\begin{table*}[!h]
\setlength{\abovecaptionskip}{0.15cm}
\caption{Test Rank@10 (\%) for multi-modal question answering with various missing and noise settings. }
\centering
\setlength\tabcolsep{3pt}   
\resizebox{0.75\textwidth}{!}{
\begin{tabular}{lc|cc|cc|cc|cc|cc|cc|cc}
\toprule
\multicolumn{1}{l}{\multirow{2}{*}{Method}} & \multirow{2}{*}{0\%} & \multicolumn{2}{c}{TMR} & \multicolumn{2}{c}{IMR} & \multicolumn{2}{c}{MMR} & \multicolumn{2}{c}{TNR} & \multicolumn{2}{c}{INR} & \multicolumn{2}{c}{TINR} & \multicolumn{2}{c}{MNR} \\[-3pt] 
\multicolumn{1}{c}{}                        &                      & 50\%       & 100\%      & 50\%       & 100\%      & 50\%       & 100\%      & 50\%       & 100\%      & 50\%       & 100\%      & 50\%        & 100\%      & 50\%       & 100\%      \\
\hline
ViLBERT                 & 74.1                              & 52.5          & 38.5          & 73.4          & 72.7          & 58.7          & 51.2          & 47.9          & 39.0            & 73.7          & 73.3          & 34.5          & 2.1           & 46.8 & 27.7 \\
ViLBERT+PKG             & 80.6                              & 71.0          & 68.9          & 79.6          & 79.1          & 73.8          & 70.9          & 70.7          & 66.3          & 80.3          & 79.9          & 69.6          & 64.5          & 71.4 & 67.3 \\
VLBERT                  & 74.6                              & 52.3          & 39.2          & 73.6          & 72.8          & 59.2          & 51.5          & 48.2          & 39.6          & 74.0          & 73.3          & 34.8          & 2.2           & 48.3 & 28.3 \\
VLBERT+PKG              & 80.9                              & 72.0          & 68.7          & 79.7          & 79.3          & 73.7          & 71.3          & 71.0          & 67.4          & 80.5          & 80.1          & 70.2          & 65.2          & 71.5 & 67.2 \\
LXMERT                  & 74.3                              & 52.1          & 38.4          & 73.5          & 72.4          & 58.4          & 50.8          & 47.4          & 39.1          & 73.6          & 73.1          & 34.6          & 2.2           & 46.8 & 27.4 \\
LXMERT+PKG              & 80.7                              & 70.9          & 68.4          & 79.8          & 78.9          & 73.6          & 71.2          & 71.0          & 66.9          & 80.2          & 79.8          & 69.5          & 64.8          & 71.2 & 66.6 \\

\hline
K3M(hard-spl)                               & \textbf{87.2}            & \textbf{79.6} & \textbf{76.8} & \textbf{86.6} & \textbf{86.3} & \textbf{81.3} & \textbf{78.9} & \textbf{79.6} & \textbf{76.5} & \textbf{86.9} & \textbf{86.6} & \textbf{77.9} & \textbf{73.7} & \textbf{79.6} & \textbf{75.3}      \\[-1pt]
\bottomrule
\end{tabular}
}

\label{T-qa}
\end{table*}
\begin{table*}[]
\setlength{\abovecaptionskip}{0.15cm}
\caption{Results of ablation for item classification (IC), product alignment (PA) and multi-modal question answering (MMQA).}

\centering
\setlength\tabcolsep{3pt}   
\resizebox{0.8\textwidth}{!}{
\begin{tabular}{clc|cc|cc|cc|cc|cc|cc|cc}
\toprule
\multirow{2}{*}{Task} & \multirow{2}{*}{Method} & \multicolumn{1}{c|}{\multirow{2}{*}{0\%}} & \multicolumn{2}{c}{TMR}       & \multicolumn{2}{c}{IMR}       & \multicolumn{2}{c}{MMR}       & \multicolumn{2}{c}{TNR}       & \multicolumn{2}{c}{INR}       & \multicolumn{2}{c}{TINR}      & \multicolumn{2}{c}{MNR} \\ [-3pt]
                      &                         & \multicolumn{1}{c|}{}                     & 50\%          & 100\%         & 50\%          & 100\%         & 50\%          & 100\%         & 50\%          & 100\%         & 50\%          & 100\%         & 50\%          & 100\%         & 50\%       & 100\%      \\
                      \hline
\multirow{4}{*}{\begin{tabular}[c]{@{}c@{}}IC\\ (accuracy \%)\end{tabular}}   & K3M w/o IFFM            & 92.4                                     & 86.9          & 83.7          & 90.3          & 87.1          & 88.4          & 84.6          & 86.8          & 83.3          & 89.2          & 86.3          & 85.7          & 81.3          & 87.6       & 83.9       \\
                      & K3M(mean)               & 93.0                                       & 87.4          & 84.3          & 90.6          & 87.4          & 88.7          & 85.3          & 87.1          & 83.7          & 89.5          & 86.7          & 85.9          & 81.2          & 88.0         & 84.3       \\
                      & K3M(soft-spl)           & 92.9                                     & 89.5          & 86.5          & 91.2          & 88.9          & 89.8          & 87.3          & 88.7          & 85.9          & 91.3          & 88.1          & 86.4          & 81.5          & 89.4       & 86.8       \\
                      & K3M(hard-spl)           & \textbf{93.2}                            & \textbf{89.9} & \textbf{86.9} & \textbf{91.7} & \textbf{89.6} & \textbf{90.2} & \textbf{87.7} & \textbf{89.2} & \textbf{86.4} & \textbf{91.6} & \textbf{88.5} & \textbf{86.8} & \textbf{81.7} & \textbf{90.1} & \textbf{87.2}       \\
                      \hline
\multirow{4}{*}{\begin{tabular}[c]{@{}c@{}}PA\\ (F1-score \%)\end{tabular}}   & K3M w/o IFFM            & 91.7                                     & 84.6          & 85.5          & 89.4          & 91.2          & 87.6          & 86.3          & 84.7          & 82.5          & 91.1          & 89.8          & 79.2          & 76.4          & 84.1       & 78.3       \\
                      & K3M(mean)               & 92.3                                     & 86.1          & 86.1          & 90.8          & 91.9          & 88.6          & 87.6          & 86.1          & 83.7          & 91.7          & 91.2          & 82.6          & 76.6          & 85.2       & 81.9       \\
                      & K3M(soft-spl)           & 92.7                                     & 86.7          & 87.1          & 91.7          & 92.4          & 89.5          & 88.5          & 87.1          & 85.9          & 92.2          & 91.7          & 84.0            & 78.2          & 87.2       & 83.1       \\
                      & K3M(hard-spl)           & \textbf{93.2}                            & \textbf{86.8} & \textbf{86.9} & \textbf{91.9} & \textbf{92.6} & \textbf{89.9} & \textbf{89.1} & \textbf{87.5} & \textbf{86.4} & \textbf{92.7} & \textbf{92.4} & \textbf{84.1} & \textbf{79.6} & \textbf{87.6} & \textbf{83.8}       \\
                      \hline
\multirow{4}{*}{\begin{tabular}[c]{@{}c@{}}MMQA\\ (Rank@10 \%)\end{tabular}} & K3M w/o IFFM            & 83.8                                     & 75.7          & 72.4          & 82.5          & 81.9          & 76.1          & 73.7          & 74.7          & 72.0            & 83.4          & 83.1          & 72.3          & 68.2          & 74.7       & 70.6       \\
                      & K3M(mean)               & 85.4                                     & 77.1          & 74.7          & 84.6          & 84.2          & 79.2          & 76.6          & 77.6          & 74.1          & 85.2          & 84.6          & 75.2          & 70.1          & 77.1       & 73.1       \\
                      & K3M(soft-spl)           & 86.5                                     & 78.6          & 75.6          & 86.0            & 85.4          & 80.9          & 77.6          & 78.2          & 75.2          & 86.2          & 85.7          & 76.6          & 71.9          & 78.1       & 74.5       \\
                      & K3M(hard-spl)           & \textbf{87.2}                            & \textbf{79.6} & \textbf{76.8} & \textbf{86.6} & \textbf{86.3} & \textbf{81.3} & \textbf{78.9} & \textbf{79.6} & \textbf{76.5} & \textbf{86.9} & \textbf{86.6} & \textbf{77.9} & \textbf{73.7} & \textbf{79.6} & \textbf{75.3}      \\[-1pt]
                      \bottomrule
\end{tabular}
}

\label{T-ablation}
\end{table*}
\subsubsection{Task definition} 
The goal of this task is to return an answer based on the multi-modal information of a given item and a question. This task can serve the automatic customer service system. For example, if a user wants to know the material or applicable season of a certain product, automatic customer service system can quickly give an answer. Following the question answering task in previous works~\cite{19DBLP:conf/iclr/SuZCLLWD20,22DBLP:conf/nips/LuBPL19,21DBLP:conf/sigir/GaoJCQLWHW20,36DBLP:journals/corr/abs-2003-13198}, we frame it as a multi-class classification task. 
Given a set of item-question pairs $\mathcal{Q}^p$ and a set of candidate answers $\mathcal{A}$, the target is to train a mapping function $f: \mathcal{Q}^p \longmapsto \mathcal{A}$.


\subsubsection{Model} 
\textbf{K3M} For a given item-question pair $(e_{ci},q_i) \in \mathcal{Q}^p$, we take $c^\ast_{qi}=M([i_c|q_i],t_c,\mathcal{TR}_c)$ as the representation of $(e_{ci},q_i)$, where $[i_c|q_i]$ is the connection of the ``question'' and the product title of the item, expressed as ``$product$ $title$ $[SEP]$ $question$''. Then we feed $c^\ast_{qi}$ into a full connection layer:
\begin{equation}
\setlength{\abovedisplayskip}{3pt}
\setlength{\belowdisplayskip}{3pt}
\label{equ_attri_gen1}
    p_i=\sigma(Wc_{qi}^\ast + \beta),
\end{equation}
where $W\in \mathbb R^{d\times |\mathcal{A}|}$ is a weighted matrix, $p_i=[p_{i1},p_{i2},...,p_{i|\mathcal{A}|}]$ where $p_{ij}$ is the probability that the answer of $(e_{ci},q_i)$ is  $a_j\in \mathcal{A}$, $j=1,...,|\mathcal{A}|$. We finetune K3M with a cross-entropy loss:
\begin{equation}
\setlength{\abovedisplayskip}{3pt}
\setlength{\belowdisplayskip}{3pt}
\label{equ_attri_gen2}
    L=-\frac{1}{|\mathcal{Q}^p|}\sum_{i=1}^{|\mathcal{Q}^p|}\sum_{j=1}^{|\mathcal{A}|} y_{ij} log(p_{ij}),
\end{equation}
where $y_{ij} = 1$ if the answer of $(e_{ci},q_i)$ is $a_j$, otherwise $y_{ij} = 0$.

\textbf{Baseline models} For ``ViLBERT'', ``LXMERT'' and ``VLBERT'', we connect the title $i_c$ and ``question''  as the text modality input, expressed as `` $product$ $title$ $[SEP]$ $question$ ''. For ``ViLBERT+PKG'', ``LXMERT+PKG'' and ``VLBERT+PKG'', we connect $i_c$, the knowledge text of $e_{ci}$, and ``question'' as the text modality input, expressed as ``$product$ $title$ $[SEP]$ $property_1$ $value_1$ $property_2$ $...$ $[SEP]$ $question$''. Following the original papers, we compute the representation of $(e_{ci},q_i)$ as the element-wise product between the last hidden states of $[CLS]$ and $[IMG]$ for ViLBERT, and as the last hidden state of $[CLS]$ for LXMERT and VLBERT. The following steps are the same as Equation~\ref{equ_attri_gen1} and ~\ref{equ_attri_gen2}.

\subsubsection{Dataset} More details are in Appendix. 



\subsubsection{Result analysis.}

The evaluation metric of this task is Rank@K (K=1, 3, 10), where Rank@K is the percentage of ground-truth answers appearing in the top-K
ranked list. In particular, same with~\cite{22DBLP:conf/nips/LuBPL19}, for a item-question pair $(e_{ci},q_i)$, we score each candidate answer $a_j \in \mathcal{A}$ as the probability that $a_j$ is its answer, that is $p_{ij}$, and then we sort all of the candidate answers in $\mathcal{A}$.

Table~\ref{T-qa}\footnote{More  results of Rank@1 and Rank@3 are shown in the Appendix.} shows the rank result of multi-modal question answering task. In this task, we can also have the similar observations as in the item classification task. 

\subsection{Ablation Study}
In this section, we verify the effectiveness of our proposal that fusing the initial features and interactive features of image and text modalities. 
We pretrain another K3M without the initial-interactive feature module (IFFM) from scratch, denoted as ``K3M w/o IFFM'', where the head entities of triples in PKG is initialized by only the interactive features of image and text modalities, and Equation~\ref{eq_initial_c} is rewritten as:

\begin{equation}
\setlength{\abovedisplayskip}{3pt}
\setlength{\belowdisplayskip}{3pt}
\label{eq_initial_c_2}
c=mean\_pooling(h_{t1}^I,...,h_{tM_1}^I,W_0h_{i1}^T,...,W_0h_{iM_2}^T).
\end{equation}

Table~\ref{T-ablation} shows the results on three downstream tasks. We can see that K3Ms with IFFM applying different fusion algorithms  all work better than K3M without IFFM, indicating that our proposed fusion of initial features and interactive features can indeed further improve the model performance by retaining the independence of text and image modalities. 
Due to limited space, we only show a part of all results, and for more  results, please refer to the Appendix.

\section{Conclusion and Future Work}
In this paper, we introduce structured knowledge of PKG into multi-modal pretraining in E-commerce, and propose a new method, K3M. The model architecture consists of modal-encoding layer for extracting the features of each modality, modal-interaction layer for modeling the interaction of multiple modalities, and modal-task layer containing different pretraining tasks for different modalities. In the modal-interaction layer, we design a structure aggregation module to propagate and aggregate the information of entity nodes and relationship edges of PKG, and design an initial-interactive feature fusion module to fuse the initial features of image and text modalities with their interactive features to further improve the model performance. Experiments on a real-world E-commerce dataset show the powerful ability of K3M. In future work, we would like to apply K3M to more downstream tasks and explore its performance on more general datasets.

\begin{acks}
This work is funded by NSFC91846204/U19B2027, national key research program 2018YFB1402800.
\end{acks}

\newpage
\normalem
\bibliographystyle{unsrt}
\bibliography{bibfile}


\appendix
\newpage
\section{Appendices}
\subsection{Additional implementation details for pretraining}
\subsubsection{Pretraining of K3M}
In experiments, the number of layers of all Transformer structures in K3M is 6. The Transformer blocks in the text encoder and the co-attention Transformer blocks of textual stream in the image-text interactor are initialized by the pretrained parameters of the 12-layer BERT-base-chinese\footnote{https://huggingface.co/bert-base-chinese} with each block having 768 hidden units and 12 attention heads. The Transformer blocks in the image encoder and the co-attention Transformer blocks of visual stream in the image-text interactor are randomly initialized with each block having 1024 hidden units and 8 attention heads. In the structure aggregation module, the representation dimension of entity and relation is set to 768 and the number of attention heads is 8.

The length of each product title is shorter than 40 tokens ($M_1$=40). The length of each object sequence is shorter than 36 ($M_2$=36) as 10 to 36 objects are extracted from each image following the previous work~\cite{22DBLP:conf/nips/LuBPL19}. The length of the knowledge text stitched by relations and tail entities is shorter than 80 ($M_3$=80). The hyper-parameter $\gamma$ of margin loss in Equation~\ref{eq_los_tri} is set at 1, and 3 negative triples are sampled for each positive triple.

We pretrained K3M with 3 different fusion algorithms used in the initial-interactive fusion module, namely ``K3M(mean)'', ``K3M(soft-spl)'' and ``K3M(hard-spl)''. They are implemented on Pytorch and trained on 8 Tsela-V100 GPUs with a total batch size of 256 for 3 epochs. We pretrain the model with Adam whose initial learning rate set at 1e-4 and use a linear decay learning rate schedule with warm up. Finally the model size is 1.7G 
and the whole training consumed 275 hours.

\subsubsection{Pretraining of baseline models}

For ``ViLBERT'', ``LXMERT'' and  ``VLBERT'', the same as K3M, we set the length of title shorter than 40 tokens, and set the length of object sequence shorter than 36.

For `ViLBERT+PKG'', ``VLBERT+PKG'' and ``LXMERT+PKG'', to introduce information of the knowledge modality (PKG) into the models, we spliced the knowledge text of PKG behind the product title and used the whole as the text modality input of the models. Specifically, we first stitch all relations and tail entities of the triples related to target item together into a long knowledge text like ``$property_1$ $value_1$ $property_2$ $value_2$ $...$''. Then we connect the product title with the knowledge text by a separator $[SEP]$ as the final text input of the baseline models, expressed as ``$product$ $title$ $[SEP]$ $property_1$ $value_1$ $property_2$ $value_2$ $...$''. The same as K3M, the length of title is shorter than 40 tokens, the length of object sequence is shorter than 36, and the length of the knowledge text shorter than 80 tokens.  

The other model settings and training details of baseline models are the same as their original papers.

\subsection{Additional implementation details and results for item classification task}
\subsubsection{Dataset}
In the dataset, there are $115,467$ items belong to $518$ classes which contain at least $50$ items used for item classification task. The number of items of train/test/dev dataset are $7:1:2$.

\subsubsection{Implementation details} We use Adam optimizer with an initial learning rate of 5e-5 and apply a linear decay learning rate schedule with warm up. We finetune the model for 4 epochs with a batch size of 32.

\subsubsection{Results}
Table~\ref{T-item_cls_m} shows the test accuracy of various models for item classification task of different modality-missing situations. Table~\ref{T-item_cls_n} shows the test accuracy of various models for item classification task of different modality-noise situations.

\begin{table*}[!h]
\setlength{\abovecaptionskip}{0.15cm}
\caption{Test Accuracy (\%) for item classification task compared with baselines with various IMRs, TMRs and MMRs. }
\label{T-item_cls_m}
\centering
\resizebox{0.8\textwidth}{!}{

\begin{tabular}{l|c|cccc|cccc|cccc}
\toprule
\multicolumn{1}{c|}{\multirow{2}{*}{Method}} & \multicolumn{1}{c|}{\multirow{2}{*}{0\%}} & \multicolumn{4}{c|}{TMR}    & \multicolumn{4}{c|}{IMR}    & \multicolumn{4}{c}{MMR}    \\
\multicolumn{1}{c|}{}                        & \multicolumn{1}{c|}{}                     & 20\% & 50\% & 80\% & 100\% & 20\% & 50\% & 80\% & 100\% & 20\% & 50\% & 80\% & 100\% \\
\hline
ViLBERT                                     &  81.8   &  73.4 & 61.8 & 54.6  & 48.9 & 80.1 & 78.7  & 75.7  & 73.8 &  76.1  & 67.0 & 62.6 & 59.4  \\
ViLBERT+PKG                                  & 92.5     & 89.7 & 86.5 & 84.4 & 83.1  & 91.8 & 89.4 & 87.6 & 86.3 & 90.4 & 88.1 & 85.5 & 84.1  \\
VLBERT                                     & 81.9          & 73.6          & 62.0            & 54.9          & 49.3          & 80.4          & 78.5          & 75.9          & 74.2          & 76.3          & 67.2          & 62.9          & 59.7       \\
VLBERT+PKG                                  & 92.6          & 89.9          & 86.8          & 84.6          & 83.4          & 91.9          & 89.6          & 87.7          & 86.4          & 90.5          & 88.3          & 85.7          & 84.4      \\
LXMERT                                & 81.6          & 73.2          & 61.7          & 54.5          & 48.5          & 80.2          & 78.4          & 75.4          & 73.6          & 76.8          & 66.7          & 62.4          & 59.3     \\
LXMERT+PKG                                  & 92.2          & 89.4          & 86.2          & 84.2          & 83.0            & 91.5          & 89.3          & 87.4          & 86.1          & 90.3          & 88.0            & 85.2          & 83.8   \\
\hline
K3M w/o IFFM                     & 92.4     & 90.4 & 86.9 & 84.8 & 83.7  & 92.2 & 90.3 & 88.2 & 87.1 & 90.8 & 88.4 & 86.0 & 84.6  \\
K3M(mean)  & 93.0     & 90.8 & 87.4 & 85.6 & 84.3  & 92.6 & 90.6 & 88.5 & 87.4 & 91.4 & 88.7 & 86.4 & 85.3  \\
K3M(soft-spl)         & 92.9     & 91.1 & 89.5 & 87.6 & 86.5  & 92.6 & 91.2 & 89.6 & 88.9 & 91.9 & 89.8 & 88.6 & 87.3  \\
K3M(hard-spl)  & \textbf{93.2}     & \textbf{91.6} & \textbf{89.9} & \textbf{88.2} & \textbf{86.9}  & \textbf{92.9} & \textbf{91.7} & \textbf{90.3} & \textbf{89.6} & \textbf{92.1} & \textbf{90.2} & \textbf{88.7} & \textbf{87.7} \\
\bottomrule
\end{tabular}

}

\end{table*}
\begin{table*}[!h]
\setlength{\abovecaptionskip}{0.15cm}
\caption{Test Accuracy (\%) for item classification task compared with baselines with various TNRs, INRs, TIMRs and MNRs.}
\label{T-item_cls_n}
\centering
\resizebox{1.0\textwidth}{!}{
\begin{tabular}{l|c|cccc|cccc|cccc|cccc}
\toprule
\multicolumn{1}{c|}{\multirow{2}{*}{Method}} & \multirow{2}{*}{0\%} & \multicolumn{4}{c|}{TNR}    & \multicolumn{4}{c|}{INR}    & \multicolumn{4}{c|}{TINR}   & \multicolumn{4}{c}{MNR}    \\
\multicolumn{1}{c|}{}                        &                   & 20\% & 50\% & 80\% & 100\% & 20\% & 50\% & 80\% & 100\% & 20\% & 50\% & 80\% & 100\% & 20\% & 50\% & 80\% & 100\% \\
\hline
ViLBERT                                     & 81.8 & 69.2 & 58.6   & 49.5 & 46.1 & 79.3  & 78.5  & 75.2& 73.6 & 65.3  & 42.7 & 17.6 & 3.7 & 70.1  & 56.8 & 46.7 & 40.8   \\
ViLBERT+PKG                                  & 92.5              & 90.4 & 86.2 & 83.5 & 82.1  & 91.4 & 89.6 & 87.1 & 85.7 & 89.1 & 84.7 & 80.8 & 78.6  & 90.5 & 87.5 & 84.9 & 83.3  \\
VLBERT                                      &   81.9        &   69.5          & 58.8          & 49.6          & 46.2          & 79.6          & 78.7          & 75.3          & 74.0            & 65.6          & 42.8          & 17.6          & 3.4           & 70.2          & 57.1          & 46.8          & 40.9          \\    
VLBERT+PKG                                &  92.6        & 90.5          & 86.5          & 83.8          & 82.2          & 91.7          & 89.9          & 87.3          & 85.9          & 89.4          & 84.8          & 80.8          & 78.5          & 90.7          & 87.7          & 85.1          & 83.4          \\
LXMERT                                      &   81.6      & 68.8          & 58.3          & 49.2          & 45.8          & 79.1          & 78.4          & 74.8          & 73.2          & 64.8          & 42.6          & 17.5          & 3.5           & 69.8          & 56.7          & 46.5          & 40.6          \\
LXMERT+PKG                                   &  92.2     & 90.1          & 85.8          & 83.2          & 81.9          & 91.1          & 89.2          & 86.9          & 85.4          & 88.7          & 84.6          & 80.2          & 78.3          & 90.3          & 87.3          & 84.6          & 83.2          \\
\hline

K3M w/o IFFM                   & 92.4              & 90.3 & 86.8 & 84.9 & 83.3  & 91.7 & 89.2 & 87.4 & 86.3   & 89.4 & 85.7 & 82.0   & 81.3  & 91.0   & 87.6 & 85.2 & 83.9  \\
K3M(mean)   & 93.0  & 90.6 & 87.1 & 85.2 & 83.7  & 91.9 & 89.5 & 87.8 & 86.7 & 89.7 & 85.9 & 81.9 & 81.2  & 91.3 & 88.0   & 85.4 & 84.3  \\
K3M(soft-spl)     & 92.9              & 90.9 & 88.7 & 86.8 & 85.9  & 92.1 & 91.3 & 89.2 & 88.1  & \textbf{90.3} & 86.4 & \textbf{82.5} & 81.5  & 91.5 & 89.4 & 88.1 & 86.8  \\
K3M(hard-spl)         & \textbf{93.2}              & \textbf{91.5} & \textbf{89.2} & \textbf{87.5} & \textbf{86.4}  & \textbf{92.8} & \textbf{91.6} & \textbf{89.7} & \textbf{88.5}  & 89.8 & \textbf{86.8} & 82.3 & \textbf{81.7}  & \textbf{91.9} & \textbf{90.1} & \textbf{88.3} & \textbf{87.2} \\
\bottomrule
\end{tabular}

}

\end{table*}

\subsection{Additional implementation details and results for product alignment task}

\subsubsection{Dataset}
In the dataset, there are $24,707$ aligned item pairs in total, and we filter out the item pairs belong to classes that contain less than $50$ aligned item pairs (two aligned items belong to the same item class in our platform). Finally, we collect $20,818$ aligned item pairs for experiment
. To generate negative item pairs, for a given pair of aligned items $(e_{ci}^0,e_{ci}^1) \in \mathcal{C}^p$, we randomly replace $e_{ci}^0$ or $e_{ci}^1$ with another item $e_{ci}'$, namely $(e_{ci}^0,e_{ci}') \notin \mathcal{C}^p$ or $(e_{ci}',e_{ci}^1) \notin \mathcal{C}^p$. We generate 3 negative item pairs for each aligned item pair in train dataset, 1 negative item pair for each aligned item pair in dev/test dataset. The number of item pairs of train/dev/test dataset are $7:1:2$.

\subsubsection{Implementation details} We use Adam optimizer with an initial learning rate of 5e-5 and apply a linear decay learning rate schedule with warm up. We finetune the model for 4 epochs with a batch size of 24.

\subsubsection{Results}
Table~\ref{T-same_m} shows the test F1-score of various models for product alignment task of different modality-missing situations. Table~\ref{T-same_n} shows the test F1-score of various models for product alignment task of different modality-noise situations.

\begin{table*}[!h]
\setlength{\abovecaptionskip}{0.15cm}
\caption{Test F1-score (\%) for product alignment task compared with baselines with various IMRs, TMRs and MMRs.}
\label{T-same_m}
\centering
\resizebox{0.8\textwidth}{!}{

\begin{tabular}{l|c|cccc|cccc|cccc}
\toprule
\multicolumn{1}{c|}{\multirow{2}{*}{Method}} & \multicolumn{1}{c|}{\multirow{2}{*}{0\%}} & \multicolumn{4}{c|}{TMR}    & \multicolumn{4}{c|}{IMR}    & \multicolumn{4}{c}{MMR}    \\
\multicolumn{1}{c|}{}                        & \multicolumn{1}{c|}{}                     & 20\% & 50\% & 80\% & 100\% & 20\% & 50\% & 80\% & 100\% & 20\% & 50\% & 80\% & 100\% \\
\hline
ViLBERT                 & 91.2                                     & 87.4          & 84.7          & 82.9          & 84.3          & 91.0          & 85.7          & 89.5          & 89.7          & 89.0          & 87.3          & 84.8          & 83.1          \\
ViLBERT+PKG             & 91.7                                     & 89.2          & 86.1          & 83.7          & 85.4          & 91.4          & 88.4          & 91.4          & 91.2          & 89.5          & 88.3          & 87.4          & 86.5          \\
VLBERT                  & 91.4                                     & 87.8          & 85.4          & 83.3          & 83.7          & 91.1          & 86.8          & 89.7          & 89.9          & 89.3          & 87.6          & 85.2          & 84.4          \\
VLBERT+PKG              & 91.8                                     & 89.1          & 86.3          & 84.8          & 85.3          & 91.5          & 89.2          & 91.6          & 91.4          & 89.6          & 88.4          & 87.1          & 86.3          \\
LXMERT                  & 91.1                                     & 86.9          & 84.5          & 82.8          & 83.3          & 90.4          & 86.1          & 88.7          & 89.1          & 88.7          & 85.7          & 84.2          & 82.8          \\
LXMERT+PKG              & 91.6                                     & 88.5          & 85.7          & 84.5          & 84.9          & 90.9          & 88.6          & 90.2          & 90.5          & 89.3          & 87.3          & 86.4          & 85.7          \\
\hline

K3M w/o IFFM   & 91.7          & 87.7          & 84.6          & 83.9          & 85.5          & 91.4          & 89.4          & 91.6          & 91.2          & 90.2          & 87.6          & 86.8
& 86.3          \\

K3M(mean)            & 92.3          & 89.8          & 86.1          & 85.3          & 86.1          & 91.8          & 90.8          & 92            & 91.9          & 90.7          & 88.6          & 87.7          & 87.6          \\

K3M(soft-spl)    & 92.7          & 88.4               & 86.7 & 85.5    & \textbf{87.1}           & 92.6          & 91.7          & 92.4          & 92.4          & 91.3          & 89.5          & 88.3          & 88.5          \\

K3M(hard-spl) & \textbf{93.2} & \textbf{90.1} & \textbf{86.8} & \textbf{86.2} &
86.9          &  \textbf{92.9} & \textbf{91.9} & \textbf{92.8} & \textbf{92.6} & \textbf{91.9} & \textbf{89.9} & \textbf{88.7} & \textbf{89.1} \\
\bottomrule
\end{tabular}

}

\end{table*}
\begin{table*}[!h]
\setlength{\abovecaptionskip}{0.15cm}
\caption{Test F1-score (\%) for product alignment task compared with baselines with various TNRs, INRs, TIMRs and MNRs.}
\label{T-same_n}
\centering
\resizebox{1.0\textwidth}{!}{
\begin{tabular}{l|c|cccc|cccc|cccc|cccc}
\toprule
\multicolumn{1}{c|}{\multirow{2}{*}{Method}} & \multirow{2}{*}{0\%} & \multicolumn{4}{c|}{TNR}    & \multicolumn{4}{c|}{INR}    & \multicolumn{4}{c|}{TINR}   & \multicolumn{4}{c}{MNR}    \\
\multicolumn{1}{c|}{}                        &                   & 20\% & 50\% & 80\% & 100\% & 20\% & 50\% & 80\% & 100\% & 20\% & 50\% & 80\% & 100\% & 20\% & 50\% & 80\% & 100\% \\
\hline
ViLBERT                 & 91.2                                     & 85.8          & 83.3          & 81.2          & 79.3          & 90.5          & 90.2          & 90.0          & 89.7          & 76.6          & 63.2          & 51.7          & 33.3          & 84.6 & 78.2 & 73.1 & 67.2 \\
ViLBERT+PKG             & 91.7                                     & 87.2          & 84.7          & 83.7          & 82.6          & 91.4          & 90.9          & 90.6          & 90.4          & 86.1          & 79.1          & 76.3          & 74.6          & 85.8 & 84.1 & 81.2 & 78.7 \\
VLBERT                  & 91.4                                     & 86.4          & 83.6          & 81.5          & 80.2          & 90.6          & 90.5          & 90.2          & 89.8          & 77.2          & 63.8          & 52.1          & 33.3          & 85.1 & 78.9 & 73.6 & 68.3 \\
VLBERT+PKG              & 91.8                                     & 87.6          & 85.1          & 84.2          & 83.4          & 91.5          & 91.1          & 90.8          & 90.6          & 86.4          & 79.3          & 77.1          & 75.1          & 86.2 & 84.6 & 81.5 & 79.7 \\
LXMERT                  & 91.1                                     & 84.6          & 83.4          & 81.0          & 79.4          & 90.3          & 89.8          & 89.3          & 89.0          & 76.2          & 62.8          & 51.6          & 33.3          & 84.7 & 77.9 & 72.7 & 67.1 \\
LXMERT+PKG              & 91.6                                     & 86.3          & 85.0          & 83.8          & 82.8          & 91.2          & 90.5          & 90.3          & 90.2          & 85.7          & 78.4          & 76.6          & 74.8          & 85.4 & 83.9 & 80.6 & 78.2 \\

\hline

K3M w/o IFFM                     & 91.7          & 87.2          & 84.7          & 83.8          & 82.5          & 91.5          & 91.1          & 90.5          & 89.8          & 86.4          & 79.2          & 76.9          & 76.4          & 86.8          & 84.1          & 81.2          & 78.3          \\
K3M(mean)    & 92.3          & 88.1          & 86.1          & 84.8          & 83.7          & 92.1          & 91.7          & 91.4          & 91.2          & 86.9          & 82.6          & 78.2          & 76.6          & 87.7          & 85.2          & 83.7          & 81.9          \\
K3M(soft-spl)      & 92.7          & 90.2          & 87.1          & 86.5          & 85.9          & 92.5          & 92.2          & 91.8          & 91.7          & \textbf{88.7} & 84            & 80.5          & 78.2          & 89.1          & 87.2          & 84.7          & 83.1          \\
K3M(hard-spl)        & \textbf{93.2} & \textbf{90.6} & \textbf{87.5} & \textbf{86.8} & \textbf{86.4} & \textbf{93.1} & \textbf{92.7} & \textbf{92.5} & \textbf{92.4} & 88.3          & \textbf{84.1} & \textbf{81.1} & \textbf{79.6} & \textbf{89.7} & \textbf{87.6} & \textbf{85.4} & \textbf{83.8} \\
\bottomrule
\end{tabular}

}

\end{table*}

\subsection{Additional implementation details and results for multi-modal question answering task}

\subsubsection{Dataset}
We generate the dataset for this task based on the $115,467$ items from the dataset of item classification task. For each item $e_c$, we randomly select and remove one triple from $\mathcal{TR}_c$, which is used to generate a question and the answer. For example, for \textit{Item-1} in Figure~\ref{fig_intro}, we remove the triple <\textit{Item-1, Material, Cotton}>, so ``What is the material of this item ?'' is the generated question and ``Cotton'' is answer. Finally, $115,467$ item-question pairs are generated, the size of the candidate answer set $\mathcal{A}$ is $4,809$.
The number of item-question pairs of train/dev/test dataset are $7:1:2$. 

\subsubsection{Implementation details} We use Adam optimizer with an initial learning rate of 5e-5 and apply a linear decay learning rate schedule with warm up. We finetune the model for 6 epochs with a batch size of 32. 

\subsubsection{Results}
Table~\ref{T-att_gen_m_rank1}, Table~\ref{T-att_gen_m_rank3} and Table~\ref{T-att_gen_m_rank10} shows the test Rank@1, Rank@3 and Rank@10 of various models for product alignment task of different modality-missing situations, respectively. Table~\ref{T-att_gen_n_rank1}, Table~\ref{T-att_gen_n_rank3} and Table~\ref{T-att_gen_n_rank10} shows the test Rank@1, Rank@3 and Rank@10 of various models for product alignment task of different modality-noise situations, respectively.
\begin{table*}[!h]
\setlength{\abovecaptionskip}{0.15cm}
\caption{Test Rank@1 (\%) for multi-modal question answer task compared with baselines with various IMRs, TMRs and MMRs.}\label{T-att_gen_m_rank1}
\centering
\resizebox{0.8\textwidth}{!}{

\begin{tabular}{l|c|cccc|cccc|cccc}
\toprule
\multicolumn{1}{c|}{\multirow{2}{*}{Method}} & \multicolumn{1}{c|}{\multirow{2}{*}{0\%}} & \multicolumn{4}{c|}{TMR}    & \multicolumn{4}{c|}{IMR}    & \multicolumn{4}{c}{MMR}    \\
\multicolumn{1}{c|}{}                        & \multicolumn{1}{c|}{}                     & 20\% & 50\% & 80\% & 100\% & 20\% & 50\% & 80\% & 100\% & 20\% & 50\% & 80\% & 100\% \\
\hline
ViLBERT                 & 57.8                                     & 48.0            & 34.9          & 23.5          & 20.3          & 57.4          & 56.8          & 56.3          & 55.9          & 51.8          & 42.5          & 37.9          & 34.3          \\
ViLBERT+PKG             & 69.1                                     & 63.2          & 58.1          & 55.6          & 55.3          & 68.4          & 68.1          & 67.5          & 67.1          & 65.2          & 61.2          & 59.7          & 57.3          \\
VLBERT                  & 58.0                                       & 47.8          & 35.2          & 23.7          & 20.6          & 57.9          & 57.3          & 56.6          & 56.3          & 52.6          & 42.8          & 38.4          & 34.9          \\
VLBERT+PKG              & 69.6                                     & 63.8          & 59.0            & 56.0            & 55.6          & 68.8          & 68.5          & 67.9          & 67.5          & 65.6          & 61.5          & 59.9          & 58.2          \\
LXMERT                  & 57.2                                     & 46.3          & 34.6          & 23.2          & 19.6          & 57.0            & 56.6          & 56.3          & 56.1          & 52.1          & 42.6          & 37.5          & 34.5          \\
LXMERT+PKG              & 68.6                                     & 62.4          & 57.4          & 55.1          & 54.9          & 68.5          & 68.3          & 67.4          & 67.3          & 65.2          & 61.2          & 59.4          & 58.0            \\
\hline
K3M w/o IFFM            & 72.1                                     & 67.3          & 60.8          & 59.1          & 58.2          & 71.7          & 71.3          & 70.7          & 70.5          & 68.5          & 62.7          & 62.6          & 59.7          \\
K3M(mean)               & 74.8                                     & 68.6          & 62.8          & 58.9          & 58.2          & 74.2          & 73.9          & 73.6          & 73.4          & 71.3          & 66.3          & 64.8          & 63.3          \\
K3M(soft-spl)           & 75.9                                     & 70.6          & 64.2          & 60.7          & 60.2          & 75.2          & 74.5          & 73.9          & 73.3          & 72.6          & 66.4          & 65.6          & 64.5          \\
K3M(hard-spl)           & \textbf{76.7}                            & \textbf{72.2} & \textbf{65.7} & \textbf{61.7} & \textbf{61.1} & \textbf{76.5} & \textbf{76.2} & \textbf{75.5} & \textbf{74.9} & \textbf{73.2} & \textbf{69.4} & \textbf{67.2} & \textbf{65.6}\\
\bottomrule
\end{tabular}
}

\end{table*}
\begin{table*}[!h]
\setlength{\abovecaptionskip}{0.15cm}
\caption{Test Rank@3 (\%) for multi-modal question answer task compared with baselines with various IMRs, TMRs and MMRs.}
\label{T-att_gen_m_rank3}
\centering
\resizebox{0.8\textwidth}{!}{

\begin{tabular}{l|c|cccc|cccc|cccc}
\toprule
\multicolumn{1}{c|}{\multirow{2}{*}{Method}} & \multicolumn{1}{c|}{\multirow{2}{*}{0\%}} & \multicolumn{4}{c|}{TMR}    & \multicolumn{4}{c|}{IMR}    & \multicolumn{4}{c}{MMR}    \\
\multicolumn{1}{c|}{}                        & \multicolumn{1}{c|}{}                     & 20\% & 50\% & 80\% & 100\% & 20\% & 50\% & 80\% & 100\% & 20\% & 50\% & 80\% & 100\% \\
\hline
ViLBERT                 & 69.2                                     & 57.4          & 44.8          & 32.7          & 29.1          & 68.6          & 68.2          & 67.8          & 67.4          & 62.1          & 52.6          & 47.1          & 43.8          \\
ViLBERT+PKG             & 76.8                                     & 70.6          & 65.6          & 63.5          & 62.7          & 76.1          & 75.6          & 75.1          & 74.7          & 72.5          & 68.6          & 67.2          & 65.3          \\
VLBERT                  & 69.5                                     & 57.8          & 44.7          & 32.9          & 29.7          & 68.9          & 68.5          & 67.9          & 67.7          & 62.6          & 52.9          & 48.3          & 44.2          \\
VLBERT+PKG              & 77.2                                     & 71.5          & 66.6          & 63.7          & 62.5          & 76.4          & 75.8          & 75.4          & 74.9          & 73.0            & 69.2          & 67.5          & 66.1          \\
LXMERT                  & 68.9                                     & 56.6          & 44.4          & 32.3          & 28.7          & 68.7          & 68.6          & 67.8          & 67.5          & 62.3          & 52.7          & 47.6          & 44.3          \\
LXMERT+PKG              & 76.7                                     & 70.0          & 65.3          & 63.1          & 62.6          & 76.3          & 75.8          & 75.3          & 75.0          & 72.9          & 68.9          & 67.3          & 65.7          \\
\hline
K3M w/o IFFM            & 79.9                                     & 74.2          & 68.5          & 66.3          & 65.1          & 79.3          & 78.9          & 78.2          & 77.6          & 75.8          & 70.6          & 71.0          & 67.8          \\
K3M(mean)               & 81.1                                     & 75.5          & 71.4          & 68.6          & 67.8          & 80.7          & 80.4          & 79.9          & 79.8          & 77.6          & 74.0          & 72.4          & 71.2          \\
K3M(soft-spl)           & 82.2                                     & 76.3          & 72.7          & 70.2          & 69.1          & 81.8          & 81.6          & 81.2          & 81.1          & 79.2          & 75.1          & 73.4          & 71.9          \\
K3M(hard-spl)           & \textbf{83.4}                            & \textbf{78.5} & \textbf{74.2} & \textbf{71.6} & \textbf{70.8} & \textbf{83.2} & \textbf{82.8} & \textbf{82.4} & \textbf{82.4} & \textbf{80.7} & \textbf{76.0} & \textbf{74.4} & \textbf{72.6}
\\
\bottomrule
\end{tabular}
}

\end{table*}
\begin{table*}[!h]
\setlength{\abovecaptionskip}{0.15cm}
\caption{Test Rank@10 (\%) for multi-modal question answer task compared with baselines with various IMRs, TMRs and MMRs.}
\label{T-att_gen_m_rank10}
\centering
\resizebox{0.8\textwidth}{!}{

\begin{tabular}{l|c|cccc|cccc|cccc}
\toprule
\multicolumn{1}{c|}{\multirow{2}{*}{Method}} & \multicolumn{1}{c|}{\multirow{2}{*}{0\%}} & \multicolumn{4}{c|}{TMR}    & \multicolumn{4}{c|}{IMR}    & \multicolumn{4}{c}{MMR}    \\
\multicolumn{1}{c|}{}                        & \multicolumn{1}{c|}{}                     & 20\% & 50\% & 80\% & 100\% & 20\% & 50\% & 80\% & 100\% & 20\% & 50\% & 80\% & 100\% \\
\hline
ViLBERT                 & 74.1                                     & 63.3 & 52.5 & 41.9 & 38.5  & 73.6 & 73.4 & 72.9 & 72.7  & 68.2 & 58.7 & 54.2 & 51.2  \\
ViLBERT+PKG             & 80.6                                     & 75.2 & 71.0 & 69.4 & 68.9  & 79.9 & 79.6 & 79.3 & 79.1  & 77.1 & 73.8 & 72.4 & 70.9  \\
VLBERT                  & 74.6                                     & 63.7 & 52.3 & 42.1 & 39.2  & 73.7 & 73.6 & 73.3 & 72.8  & 68.1 & 59.2 & 55.1 & 51.5  \\
VLBERT+PKG              & 80.9                                     & 76.0 & 72.0 & 69.1 & 68.7  & 80.1 & 79.7 & 79.4 & 79.3  & 77.0 & 73.7 & 72.6 & 71.3  \\
LXMERT                  & 74.3                                     & 62.8 & 52.1 & 41.6 & 38.4  & 73.6 & 73.5 & 73.2 & 72.4  & 67.6 & 58.4 & 54.3 & 50.8  \\
LXMERT+PKG              & 80.7                                     & 74.6 & 70.9 & 69.2 & 68.4  & 79.9 & 79.8 & 79.4 & 78.9  & 77.5 & 73.6 & 72.5 & 71.2  \\
\hline
K3M w/o IFFM            & 83.8                                     & 79.2 & 75.7 & 73.4 & 72.4  & 82.8 & 82.5 & 82.1 & 81.9  & 80.2 & 76.1 & 75.0 & 73.7  \\
K3M(mean)               & 85.4                                     & 81.4 & 77.1 & 75.8 & 74.7  & 84.9 & 84.6 & 84.5 & 84.2  & 82.5 & 79.2 & 77.9 & 76.6  \\
K3M(soft-spl)           & 86.5                                     & 82.3 & 78.6 & 77.2 & 75.6  & 86.1 & 86.0 & 85.7 & 85.4  & 84.1 & 80.9 & 79.1 & 77.6  \\
K3M(hard-spl)           & \textbf{87.2}                            & \textbf{83.4} & \textbf{79.6} & \textbf{78.1} & \textbf{76.8} & \textbf{86.8} & \textbf{86.6} & \textbf{86.4} & \textbf{86.3} & \textbf{84.4} & \textbf{81.3} & \textbf{80.1} & \textbf{78.9} 
\\
\bottomrule
\end{tabular}

}

\end{table*}

\begin{table*}[!h]
\setlength{\abovecaptionskip}{0.15cm}
\caption{Test Rank@1 (\%) for multi-modal question answer task compared with baselines with various INRs, TNRs, TINRs and MNRs.}
\label{T-att_gen_n_rank1}
\centering
\resizebox{1.0\textwidth}{!}{
\begin{tabular}{l|c|cccc|cccc|cccc|cccc}
\toprule
\multicolumn{1}{c|}{\multirow{2}{*}{Method}} & \multicolumn{1}{c|}{\multirow{2}{*}{0\%}} & \multicolumn{4}{c|}{TNR}                                       & \multicolumn{4}{c|}{INR}                                       & \multicolumn{4}{c|}{TINR}                                      & \multicolumn{4}{c}{MNR}                                       \\
                        & \multicolumn{1}{c|}{}                     & 20\%          & 50\%          & 80\%          & 100\%         & 20\%          & 50\%          & 80\%          & 100\%         & 20\%          & 50\%          & 80\%          & 100\%         & 20\%          & 50\%          & 80\%          & 100\%         \\
                        \hline
ViLBERT                 & 57.8                                     & 44.7          & 30.6          & 20.8          & 20.2          & 57.4          & 57.1          & 56.6          & 56.4          & 45            & 26.1          & 8.3           & 0.2           & 47.9          & 33.7          & 22.6          & 17.5          \\
ViLBERT+PKG             & 69.1                                     & 61.8          & 57.7          & 54.1          & 53.2          & 68.9          & 68.7          & 68.4          & 68.2          & 62.4          & 55.8          & 49.9          & 48.6          & 63.6          & 58.3          & 53.3          & 52.5          \\
VLBERT                  & 58.0                                     & 45.5          & 30.8          & 21.1          & 20.4          & 57.8          & 57.4          & 57.1          & 56.8          & 45.4          & 26.3          & 8.6           & 0.2           & 48.5          & 34.8          & 22.8          & 18            \\
VLBERT+PKG              & 69.6                                     & 62.7          & 57.6          & 55.4          & 54            & 69.1          & 69.3          & 68.8          & 68.6          & 63.2          & 56.3          & 50.2          & 49.6          & 63.8          & 58.6          & 53.7          & 52.7          \\
LXMERT                  & 57.2                                     & 45.2          & 30.6          & 20.7          & 20.3          & 56.9          & 56.7          & 56.3          & 55.7          & 45.3          & 26.1          & 8.2           & 0.2           & 48.0          & 33.3          & 22.3          & 16.9          \\
LXMERT+PKG              & 68.6                                     & 62.2          & 57            & 54.9          & 53.6          & 68.4          & 68.0          & 67.7          & 67.4          & 62.5          & 55.7          & 48.2          & 48.9          & 62.8          & 58.0          & 52.8          & 52.3          \\
\hline
K3M w/o IFFM            & 72.1                                     & 65.2          & 57.2          & 58.3          & 57.1          & 71.5          & 71.2          & 70.7          & 70.3          & 64.7          & 57.5          & 49.3          & 52.5          & 66.9          & 61.2          & 55.8          & 53.8          \\
K3M(mean)               & 74.8                                     & 67.6          & 61.8          & 59.3          & 58.4          & 74.0          & 74.2          & 73.8          & 73.4          & 67.4          & 58.3          & 49.6          & 51.9          & 69.6          & 64.4          & 58.9          & 57.6          \\
K3M(soft-spl)           & 75.9                                     & 68.3          & 63.2          & 60.8          & \textbf{60.6} & 75.5          & 75.1          & 74.8          & 74.5          & 67.4          & 59.9          & 55.1          & 54.7          & 70.1          & 65.1          & 61.1          & 59.1          \\
K3M(hard-spl)           & \textbf{76.7}                            & \textbf{70.4} & \textbf{64.4} & \textbf{61.1} & 60.4          & \textbf{76.2} & \textbf{75.9} & \textbf{75.5} & \textbf{75.3} & \textbf{70.1} & \textbf{62.2} & \textbf{56.9} & \textbf{55.3} & \textbf{71.2} & \textbf{66.7} & \textbf{61.6} & \textbf{59.8}
\\

\bottomrule
\end{tabular}

}

\end{table*}
\begin{table*}[!h]
\setlength{\abovecaptionskip}{0.15cm}
\caption{Test Rank@3 (\%) for multi-modal question answer task compared with baselines with various INRs, TNRs, TINRs and MNRs.}
\label{T-att_gen_n_rank3}
\centering
\resizebox{1.0\textwidth}{!}{
\begin{tabular}{l|c|cccc|cccc|cccc|cccc}
\toprule
\multicolumn{1}{c|}{\multirow{2}{*}{Method}} & \multicolumn{1}{c|}{\multirow{2}{*}{0\%}} & \multicolumn{4}{c|}{TNR}                                       & \multicolumn{4}{c|}{INR}                                       & \multicolumn{4}{c|}{TINR}                                      & \multicolumn{4}{c}{MNR}                                       \\
                        & \multicolumn{1}{c|}{}                     & 20\%          & 50\%          & 80\%          & 100\%         & 20\%          & 50\%          & 80\%          & 100\%         & 20\%          & 50\%          & 80\%          & 100\%         & 20\%          & 50\%          & 80\%          & 100\%         \\
                        \hline
ViLBERT                 & 69.2                                     & 54.5          & 40.0          & 30.3          & 29.8          & 69.0          & 68.6          & 68.3          & 67.9          & 53.2          & 31.0          & 10.2          & 0.6           & 57.4          & 41.5          & 28.9          & 22.9          \\
ViLBERT+PKG             & 76.8                                     & 69.1          & 65.2          & 62.3          & 61.2          & 76.5          & 76.2          & 75.8          & 75.5          & 69.8          & 63.5          & 58.9          & 57.7          & 70.8          & 66.2          & 61.2          & 60.7          \\
VLBERT                  & 69.5                                     & 54.8          & 40.4          & 30.8          & 30.1          & 69.3          & 69.1          & 68.6          & 68.4          & 53.9          & 31.4          & 10.5          & 0.6           & 57.6          & 42.7          & 29.3          & 23.5          \\
VLBERT+PKG              & 77.2                                     & 70.3          & 65.5          & 63.4          & 61.9          & 76.8          & 76.5          & 76.1          & 75.7          & 70.2          & 64.1          & 59.2          & 58.4          & 71.1          & 66.4          & 61.3          & 61.1          \\
LXMERT                  & 68.9                                     & 54.7          & 39.7          & 30.5          & 30.0          & 68.7          & 68.3          & 67.9          & 67.7          & 53.5          & 31.2          & 10.1          & 0.6           & 57.1          & 41.1          & 28.8          & 22.6          \\
LXMERT+PKG              & 76.7                                     & 69.6          & 65.1          & 63.2          & 61.6          & 76.5          & 76.2          & 75.7          & 75.5          & 70.0          & 63.4          & 58.2          & 57.3          & 70.3          & 65.7          & 60.8          & 60.0          \\
\hline
K3M w/o IFFM            & 79.9                                     & 73.2          & 65.9          & 67.4          & 66.2          & 79.5          & 79.1          & 78.7          & 78.4          & 72.5          & 66.2          & 62.1          & 60.5          & 74.4          & 69.1          & 64.1          & 63.3          \\
K3M(mean)               & 81.1                                     & 75.8          & 70.8          & 68.4          & 67.4          & 80.9          & 80.7          & 80.4          & 80.3          & 75.9          & 67.3          & 64.3          & 63.2          & 75.2          & 71.4          & 66.2          & 65.4          \\
K3M(soft-spl)           & 82.2                                     & 76.5          & 71.3          & 69.8          & 69.3          & 81.8          & 81.6          & 81.4          & 81.3          & 76.2          & 68.9          & 65.8          & 64.1          & 76.7          & 72.3          & 69.1          & 67.2          \\
K3M(hard-spl)           & \textbf{83.4}                            & \textbf{77.6} & \textbf{72.1} & \textbf{68.9} & \textbf{68.5} & \textbf{83.2} & \textbf{82.9} & \textbf{82.7} & \textbf{82.6} & \textbf{77.4} & \textbf{70.2} & \textbf{66.2} & \textbf{64.8} & \textbf{77.8} & \textbf{73.2} & \textbf{69.9} & \textbf{68.6}
\\

\bottomrule
\end{tabular}

}

\end{table*}
\begin{table*}[!h]
\setlength{\abovecaptionskip}{0.15cm}
\caption{Test Rank@10 (\%) for multi-modal question answer task compared with baselines with various INRs, TNRs, TINRs and MNRs.}
\label{T-att_gen_n_rank10}
\centering
\resizebox{1.0\textwidth}{!}{
\begin{tabular}{l|c|cccc|cccc|cccc|cccc}
\toprule
\multicolumn{1}{c|}{\multirow{2}{*}{Method}} & \multicolumn{1}{c|}{\multirow{2}{*}{0\%}} & \multicolumn{4}{c|}{TNR}                                       & \multicolumn{4}{c|}{INR}                                       & \multicolumn{4}{c|}{TINR}                                      & \multicolumn{4}{c}{MNR}                                       \\
                        & \multicolumn{1}{c|}{}                     & 20\%          & 50\%          & 80\%          & 100\%         & 20\%          & 50\%          & 80\%          & 100\%         & 20\%          & 50\%          & 80\%          & 100\%         & 20\%          & 50\%          & 80\%          & 100\%         \\
                        \hline
ViLBERT                 & 74.1                                     & 60.9          & 47.9          & 39.9          & 39.0          & 73.8          & 73.7          & 73.4          & 73.3          & 57.6          & 34.5          & 12.6          & 2.1           & 62.7          & 46.8          & 34.1          & 27.7          \\
ViLBERT+PKG             & 80.6                                     & 74.1          & 70.7          & 68.2          & 66.3          & 80.4          & 80.3          & 80.1          & 79.9          & 74.6          & 69.6          & 66.5          & 64.5          & 75.3          & 71.4          & 67.3          & 67.3          \\
VLBERT                  & 74.6                                     & 61.2          & 48.2          & 40.1          & 39.6          & 74.3          & 74.0          & 73.6          & 73.3          & 58.1          & 34.8          & 12.8          & 2.2           & 62.9          & 48.3          & 34.2          & 28.3          \\
VLBERT+PKG              & 80.9                                     & 74.9          & 71.0          & 69.1          & 67.4          & 80.8          & 80.5          & 80.2          & 80.1          & 75.1          & 70.2          & 66.2          & 65.2          & 75.5          & 71.5          & 67.7          & 67.2          \\
LXMERT                  & 74.3                                     & 61.0          & 47.4          & 40.3          & 39.1          & 73.8          & 73.6          & 73.2          & 73.1          & 58.0          & 34.6          & 12.4          & 2.2           & 62.4          & 46.8          & 33.1          & 27.4          \\
LXMERT+PKG              & 80.7                                     & 74.4          & 71.0          & 68.6          & 66.9          & 80.5          & 80.2          & 80.1          & 79.8          & 74.3          & 69.5          & 64.7          & 64.8          & 74.6          & 71.2          & 67.1          & 66.6          \\
\hline
K3M w/o IFFM            & 83.8                                     & 77.3          & 74.7          & 73.5          & 72.0          & 83.6          & 83.4          & 83.2          & 83.1          & 77.8          & 72.3          & 69.5          & 68.2          & 78.7          & 74.7          & 71.9          & 70.6          \\
K3M(mean)               & 85.4                                     & 80.7          & 77.6          & 75.1          & 74.1          & 85.3          & 85.2          & 84.7          & 84.6          & 80.2          & 75.2          & 71.2          & 70.1          & 77.8          & 77.1          & 74.4          & 73.1          \\
K3M(soft-spl)           & 86.5                                     & 80.7          & 78.2          & 75.6          & 75.2          & 86.3          & 86.2          & 85.9          & 85.7          & 80.3          & 76.6          & 72.8          & 71.9          & 81.7          & 78.1          & 75.2          & 74.5          \\
K3M(hard-spl)           & \textbf{87.2}                            & \textbf{81.8} & \textbf{79.6} & \textbf{77.0} & \textbf{76.5} & \textbf{87.1} & \textbf{86.9} & \textbf{86.7} & \textbf{86.6} & \textbf{81.9} & \textbf{77.9} & \textbf{74.3} & \textbf{73.7} & \textbf{82.1} & \textbf{79.6} & \textbf{76.4} & \textbf{75.3}
 \\

\bottomrule
\end{tabular}

}

\end{table*}


\end{document}